\documentclass[11pt]{article}
\usepackage[protrusion=true,expansion=true]{microtype}
\usepackage[text={6.75in,9.2in}, top=0.8in, left=1in]{geometry}
\usepackage[table,x11names,dvipsnames,svgnames]{xcolor}
\usepackage[noindentafter,compact]{titlesec}
\usepackage[hyphens,spaces]{url}
\usepackage{amsmath}
\titlespacing{\section}{0pt}{3.5ex plus 1ex minus 0.2ex}{0.8ex plus 0.2ex}
\titlespacing{\subsection}{0pt}{3.5ex plus 1ex minus 0.2ex}{0.8ex plus 0.2ex}
\titlespacing{\subsubsection}{0pt}{3.5ex plus 1ex minus 0.2ex}{0.8ex plus 0.2ex}
\titlespacing{\paragraph}{0pt}{3.5ex plus 1ex minus 0.2ex}{0.8ex plus 0.2ex}
\titlespacing{\subparagraph}{0pt}{3.5ex plus 1ex minus 0.2ex}{0.8ex plus 0.2ex}
\titleformat{\section}{\normalfont\bfseries}{\thesection}{0.5em}{}
\titleformat{\subsection}{\normalfont\bfseries}{\thesubsection}{0.5em}{}
\titleformat{\subsubsection}{\normalfont\bfseries}{\thesubsubsection}{0.5em}{}
\titleformat{\paragraph}{\normalfont\bfseries}{\theparagraph}{0.5em}{}
\titleformat{\paragraph}{\normalfont\bfseries}{\thesubparagraph}{0.5em}{}
\usepackage{setspace}
\usepackage{xspace}                  
\usepackage{amssymb,amsfonts,amsmath,amsthm,eucal,mathtools,bigints}
\usepackage{esdiff}                  
\usepackage[T1]{fontenc}              
\usepackage{ucs}                      
\usepackage[utf8]{inputenc}
\usepackage{mathptmx}
\usepackage{nccmath}                     
\usepackage{graphicx}
\usepackage{longtable}                 
\usepackage{array}                     

\usepackage{graphicx}
\usepackage[utf8]{inputenc}
\usepackage{listings}
\usepackage{fancyvrb}
\usepackage{latexsym,amsmath,amssymb} 
\input{psfig.sty}
\usepackage{graphicx}

\title{Genetic Programming visitation scheduling solution can deliver a less austere COVID-19 pandemic population lockdown}
\author{Daniel Howard\footnote{Howard Science Limited, UK, dr.daniel.howard@gmail.com. Past Company Fellow and Capability Leader in AI, QinetiQ Group PLC (formerly UK MOD's Defence Evaluation and Research Agency). Member of the Senior Common Room, Pembroke College, Oxford.}}

\begin{document}

\maketitle
\thispagestyle{empty}   
\begin{abstract}
A computational methodology is introduced to minimize infection opportunities for people suffering some degree of lockdown in response to a pandemic, as is the 2020 COVID-19 pandemic.  Persons use their mobile phone or computational device to request trips to places of their need or interest indicating a rough time of day: `morning', `afternoon', `night' or `any time' when they would like to undertake these outings as well as the desired place to visit.  An artificial intelligence methodology which is a variant of Genetic Programming studies all requests and responds with specific time allocations for such visits that minimize the overall risks of infection, hospitalization and death of people.  A number of alternatives for this computation are presented and results of numerical experiments involving over 230 people of various ages and background health levels in over 1700 visits that take place over three consecutive days.  A novel partial infection model is introduced to discuss these proof of concept solutions which are compared to  round robin uninformed time scheduling for visits to places.  The computations indicate vast improvements with far fewer dead and hospitalized. These auger well for a more realistic study using accurate infection models with the view to test deployment in the real world.  The input that drives the infection model is the degree of infection by taxonomic class, such as the information that may arise from population testing for COVID-19 or, alternatively, any contamination model.  The taxonomy class assumed in the computations is the likely level of infection by age group. 
\end{abstract}

\section{Motivation}

The {\it quaranta giorni} or forty day isolation by the Venetians, as the name implies, was a measure applied to incoming ships~\cite{venice} which evolved into containment practices to handle recurrent epidemics\footnote{why 40 days? because the bubonic plague had a 37-day period from infection to death}. At the time of writing, owing to ubiquitous world travel, COVID-19 `quarantines' or lockdowns keep millions of people around the world confined mostly to the home for months before an `easing of measures' gradually re-opens society.  

The 2020 lockdowns in response to the COVID-19 pandemic enforce in different ways. Argentine and Spanish lockdowns are strictly policed with citizens required to make written application for outings.  In contrast, in north western Europe and the United States, lockdows are not as strict, with Denmark and the United Kingdom entrusting their citizens not to infringe lockdown rules, and liberal Sweden choosing not to lock down in an official sense but instead practicing a small number of restrictive measures. 

The manner of lockdown and how to exit a lockdown are problems that are short of informed solutions.  It is useful to discuss the generic lockdown problem as requiring an innovation capable of overcoming the following trade-off: {\it generally, the longer and more extensive is a lockdown the more effective becomes society's ability to prepare hospital facilities to control the spread of the disease but the higher are important negative factors: loss of personal freedoms; damage to the economy; poorer personal psychology; social unrest; abusive relations; undetected crime; higher incidence of other disease because people are too scared to visit the emergency room or doctor and negative effects on care of both vulnerable and elderly.}

\section{Solution}
\begin{figure}
\begin{centering}
\includegraphics[width=12.0cm]{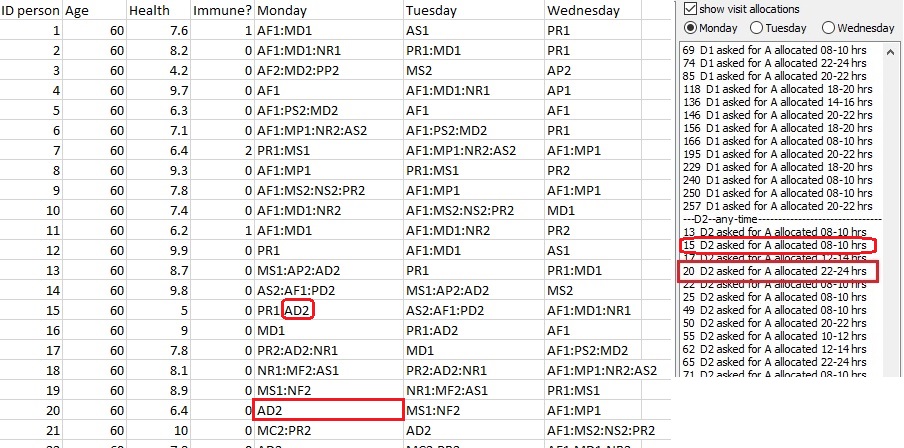}\\
\caption{(left) A part of the data file used in this work as explained in the text; (right) the solution.} \label{fig:datafile}
\end{centering}
\end{figure}
The idea is to enable lockdown whilst minimizing many of its negative consequences, e.g., to personal psychology, economics and health.  Indeed, it would be nice if life could continue as normal while in lockdown, a seemingly contradictory statement.  Inspiration comes from a crude approach taken by governments to ensure that those who venture outdoors are fewer in number. Panama~\cite{panama} used the last number of an identity document to assign two hour time slots to venture away from the home for essentials and, as Spain eased measures, it allowed  certain age groups to go out at different times of the day.  The open literature also explores changing a general lockdown to a number of partial ones~\cite{delockdown}.

The solution presented here is for citizens in lockdown to enter into smart-phone, handset, tablet or computer, a schedule of the places that they wish to visit on that day, or future dates, together with a rough idea of the part of the day they would prefer for such outings to take place. A method of optimization, in this proof of concept this is a Genetic Programming~\cite{koza:book} method, takes these requests and simulates the outings by means of an infection model, to discover a nearly optimal allocation of precise time slots for visits that reduce the likely hospitalization and death numbers. It then communicates the time allocations to citizens on their devices such that they will carry out the journeys more safely than at any times of their choice.  The solution is proactive.

Specifically, it is useful to discuss the problem within the implementation of this proof of concept.  A data file captures all requests as shown on the left side of Figure~\ref{fig:datafile}.  The visit requests on each of three consecutve days can be many, and each is denoted by a three symbol key. Requests are separated by the colon symbol.  The first letter concerns a broad time of day requested for the visit limited to:
\begin{flalign}
&  \mbox{M  wishes for the visit to take place during `morning' hours; } \nonumber \\
&  \mbox{P  wishes for the visit to take place during `afternoon' {\em post meridiem} hours; } \nonumber \\
&  \mbox{N  wishes for the visit to take place during `night' hours; } \nonumber \\
&  \mbox{A  anytime: does not mind at what hour on this day. } \nonumber
\end{flalign}
\noindent and the second letter is the desired place of visit, limited here to six types of establishment:
\begin{flalign}
&  \mbox{F  may represent a `supermarket' selling food; } \nonumber\\
&  \mbox{C  may represent a sports `club'; } \nonumber \\
&  \mbox{P  could represent a `park'; } \nonumber \\
&  \mbox{D  may stand for `doctor' a doctor's surgery centre; } \nonumber \\
&  \mbox{R  may stand for `restaurant'; } \nonumber \\
&  \mbox{S  could represent a `social' establishment; } \nonumber
\end{flalign}
\noindent the third symbol can be 1 or 2, because there are only two of each type of establishment, or a total of 12 establishments available for visits.

For example, one such request is from Person ID 20 who wants to go to Doctor's Surgery 2 at any time of day.  The day is divided into eight two hour slots. After analysis of all request, by working with an infection model, the computer generated optimization minimizes the total risk of COVID-19 infection, hospitalization and death. It allocates the time slots to the requests and the solution is communicated back to citizens. In this case Person ID 20 is allocated a time slot within the constraint that they imposed as shown on the right side of Figure~\ref{fig:datafile}.  There are other data fields such as age and health in the left side of the figure that are discussed further in this text.

\section{Data set for the proof of concept numerical experiments}
\label{section:data}

The proof of concept simulations cover three consecutive days denoted as:  Monday, Tuesday and Wednesday, involving 1704 visits as requested by 282 people.  A day consists of eight two-hour visitation slot periods available to schedule the visits.  Details of the data for the purpose of the approximate reproduction of results is as in Table~\ref{tab:data2}.  The data is an entire fabrication but common sense governed its choices: older people carry out relatively fewer outings than the young, and are in a poorer state of health.  

The degree of health of a person is a number that ranges from 1 to 10.  In any future real-world application of this research, the general health of a person, which is a measure of their immune system response to the pandemic and assumed to abate the probability of hospitalization or death could be gathered from patient records.  As seen in later sections, this level of health combines with infection and plays a pivotal role in determining which solutions are better than others. The idea is to reduce hospitalizations and fatalities.

\renewcommand{\baselinestretch}{1.4} 
\begin{table}[t!]
\begin{center}
\begin{tabular}[scale=0.55]{ | r | r | r | r | r | r | r | r | r |}
\hline
age            		& 20     & 30     & 40     & 50     & 60     & 70     & 80     & total\\ \hline
number of people	& 35     & 65     & 49     & 43     & 27     & 43     & 20     & 282 \\ \hline
average health 	&  9.49 & 9.08  & 8.51  & 7.27  & 7.86  &  5.45& 4.10&  \\
minimum health        &  9      & 8.1    &   7     & 5.1    & 4.2    &  2.1  & 1.3  &  \\
maximum health       & 10     &10      &  9.9  & 9.9    & 10     &   9    &  7   &  \\
variance health        & 0.09 & 0.29  & 0.65 & 2.37  & 2.29  &3.31 & 3.69 &  \\ \hline
Monday visits  & & & & & & &  &  586\\
average                   & 2.74  & 2.08  & 2.41  & 2.23  & 2.26 &   1.30& 1.2 &  \\
minimum                   & 1       &  1       & 1       & 1  &     1   &    1    &     1&  \\
maximum                  & 5       &  5       & 4       & 4  &    4    &    3    &      3   &  \\
variance                     & 0.53  &  0.69  & 0.65  & 0.46  &  0.85 &  0.30&0.26&  \\ \hline
Tuesday visits  & & & & & & &  &  572\\
average                   &  2.57 & 2.06  & 2.37  & 2.23 & 2.19 &1.26 &  1.15 &  \\
minimum                   &  1      &  1      &  1      & 1      &    1   &    1   &      1&  \\
maximum                  &   4     &  5      &  4     & 4       &    4   &    3  &      2  &  \\
variance                     &  0.53 &  0.67 & 0.68 & 0.50 &  0.89&0.24 &  0.13  &  \\\hline 
Wednesday visits  & & & & & & &  &  546\\
average                   &   2.6  &  1.88 & 2.06  & 2.12 & 1.85 &  1.30&   1.75 &  \\
minimum                   &    1    &  1      &  1      & 1   &   1    &   1     &      1 &  \\
maximum                  &    4    &  4      &  3     & 3  &    4   &   3    &        4&  \\
variance                     &  0.47 & 0.54  & 0.79 & 0.47   &  0.94&0.40  &0.69  &  \\
\hline
\end{tabular}
\end{center}
\caption{282 people over three consecutive days undertake a total of 1704 visits to places of their interest during lockdown. Statistical measures on health levels and visits per day per age group are shown.}
\label{tab:data2}
\end{table}

\section{Infection models used in this proof of concept}

The solution must simulate the infection process cumulatively and longitudinally in time as people go from place to place to optimize the allocation of time slots. This requires as component a model of COVID-19 infection.  The proof of concept develops a model based on person to person transmission based on simple probabilities of meetings between people in a confined location and is necessarily a very basic model but one that illustrates the potential of the solution.  

Two models are presented: a partial infection probability developed for this work and a simple standard probability model.  Both are simple and would need to be improved by epidemiologists for their real world application.

\subsection{A partial infection probability model}

The idea of a `partially infected' individual is developed here because it will only be possible in some average sense to know who might be infected\footnote{if we knew precisely who was infected they could be isolated and the solution becomes unnecessary}.  The idea is to assign a `partial infection level' to all members of a taxonomy class of interest, and then simulate how they will infect others and further infect each other.

If an infected individual $I$ is in close proximity to a susceptible individual $S$ then the possibility exists of transmission of the disease from $I$ to $S$. Without loss of generality the assumption is made that every such encounter will result in infection transmission. An infection probability based on a count of such encounters between $S$ and one or more $I$ can be expressed as $p_n$ where $n$ is the number of infected who may come into contact with susceptible in a fixed interval of time that denotes the duration of a visit, e.g., an hour or two.  More sophisticated relations should consider this time dimension and may model it with Poisson distributions but such complications are ignored for the purpose of this presentation. 

If the location of the encounters is for example, a store of a certain physical size, and $s$ is the number of sub-locations of that are available to visitors that together comprise the walkable area of that store, they are sub-units of area small enough such that people may position themselves and come into close proximity of each other, then a simple count of probability for such encounters between $S$ and one or more $I$ leads to:  $p_n = 1 - ((s-1)/s)^n$. A property of this relation is the convergence to one as $n$ grows large: $n \to\infty$ then $p_\infty \to 1.0$ meaning that a non-infected person will surely come into contact with one or more infected at that store. Moreover, also $p_2 < 2 p_1$ such that if doubling $I$ then the probability of infection for $S$ is increased but never doubled.

In driving a contact based infection model, simulations must assume a certain number of persons are infected {\em a priori} at the start of the simulation.  This presents a challenge because it requires assumptions of who is infected and why, and also a huge number of stochastic computation and an averaging of results.  As each person is different and not a clear member of a taxonomy class\footnote{the idea will not be unfamiliar to mathematical biologists and epidemiologists~\cite{taxonomy}}, for example, people may be of the same age but of different health level granting them less resistance to the infection, the computations would necessitate carefully balancing assumptions about who should be assumed to be infected to drive the computations. 

The partial infection model that is presented, while not being itself a perfect solution, does not have this onerous requirement and simplifies the computational effort requirement for this proof of concept study because it can take an idea about how infected are members of the population, assigning a probability of infection to all members of that taxonomy grouping in order to drive the infection simulation.

The concept of a partially infected individual is a modelling tool.  Each person $P_j$ is represented by a vector of size two, $P_j = (S_j, I_j)$ with $S_j + I_j = 1$.  For example a person with ID label $j=1$ that is forty percent infected is represented by $S_1 = 0.6$ and $I_1=0.4$ and another with ID label $j=102$ who is one percent infected by $S_{102} = 0.99$ and $I_{102} = 0.01$.  

Consider the meeting at the store of $n_p$ persons of which $n$ have some degree of partial infection.  The resulting infection pressure,  $pI_n$, that is the resultant partial infection probability of the encounter that is  brought about by the infection contributions of the $n$ partially or fully infected persons, can be obtained for two cases:  

$S_{max}$ is the maximum $S_j$ out of all $n$ persons.  With the exception of the owner of $S_{max}$  each $I_k$ of the other partially or fully infected persons is multiplied by $S_{max}$ and by a multiplicative numerical constants $g_j$ soon to be discussed.  These product terms are summed to obtain the probability of infection for the encounter. When $n_p=n$, the number of such product terms is therefore $n-1$.  However, when $n_p>n$, $S_{max}$ is considered to be from one of the fully susceptible persons in $n_p$ and thus $S_{max}=1$. Now the number of such product terms in the sum is $n$. A formula to compute $pI_n$ given $n_p=n$ with every participant infected or partially infected, is developed as
\begin{align}
pI_n = \sum_{j=1}^{n-1} S_{max} I_j   g_j\hspace{0.2cm}\mbox{ where}\hspace{0.3cm}  g_j = \left( \frac{1}{s}\right) \left(\frac{s-1}{s}\right)^{j-1} \hspace{0.2cm}\mbox{and}\hspace{0.3cm} I_1 > I_2 > ... > I_{n-2}>I_{n-1}, 
\end{align}
\noindent and for the case  $n_p>n$ with $S_{max}=1$ as
\begin{align}
pI_n = \sum_{j=1}^{n} I_j  g_j \hspace{0.2cm}\mbox{ where}\hspace{0.3cm}  g_j = \left( \frac{1}{s}\right) \left(\frac{s-1}{s}\right)^{j-1} \hspace{0.2cm}\mbox{and}\hspace{0.3cm} I_1 > I_2 > ... > I_{n-1}>I_{n}.
\end{align}
the constants $g_j$ emanate from simple overlap counts in probability trees as in appendix~\ref{sec:partial} and this author's earlier technical communication~\cite{inha2020}. Deliberately the products are arranged or ordered so that the largest $g_j$ corresponds to the largest $I_j$.  This represents the worst infection case which subsumes all others.

Once the probability of infection $pI_n$ is calculated, the partial infection of all participants in $n_p$ is updated for encounter $v$ in readiness for the next visit $v+1$  as follows:
\begin{align}
I_j^{v+1} = pI_n S_j^{v} + I_j^v, \hspace{1cm} S^{v+1} = 1.0 -  I^{v+1} \hspace{0.6cm}\mbox{for all}\hspace{0.2cm} j=1,...,n_p.
\end{align}
\subsubsection{A worked out example}

As an example, assume $s=4$ and $n_p=n=4$. The four participants are partially infected as follows,
\begin{flalign}
&  \mbox{P1 has I}=0.01 \mbox{ and S}=0.99, \nonumber \\
&  \mbox{P2 has I}=0.98 \mbox{ and S}=0.02, \nonumber \\
&  \mbox{P3 has I}=0.97 \mbox{ and S}=0.03, \nonumber \\
&  \mbox{P4 has I}=0.99 \mbox{ and S}=0.01. \nonumber 
\end{flalign}
\noindent which approaches the situation of three fully infected and one susceptible. Participant $P1$ has the highest component of S, thus $S_{max}= 0.99$.  The probability of infection is computed as
\begin{flalign}
&pI_n=[ (0.99)(0.99)(16) + (0.99)(0.98)(12) + (0.99)(0.97)(9) ]/64 = 0.5708,\nonumber
\end{flalign}
\noindent This number $pI_n = 0.5708$ is very close to  $p_3 = 1 - (3/4)^3 = 0.5781$. It is smaller because the infected do not contribute a complete level of infection.  The updated partial infections for all four participants in readiness for subsequent encounters are:
\begin{flalign}
&  \mbox{P1 has I}=(0.99)(0.5708) + 0.01 = 0.5751 \mbox{ and S}= 0.4249,  \nonumber \\
&  \mbox{P2 has I}=(0.02)(0.5708) + 0.98 = 0.9914 \mbox{ and S}= 0.0086,  \nonumber \\
&  \mbox{P3 has I}=(0.03)(0.5708) + 0.97 = 0.9871 \mbox{ and S}= 0.0129,  \nonumber \\
&  \mbox{P4 has I}=(0.01)(0.5708) + 0.99 = 0.9957 \mbox{ and S}= 0.0043.  \nonumber
\end{flalign}

\subsubsection{A second worked out example}
A second example with $s=6$ has seven participants $n_p = 7$ but only three are infected $n=3$. In this case, $n_p>n$ and $S_{max} = 1.0$. Assume the infected are as follows:
 \begin{flalign}
&  \mbox{P1 has I}=0.3 \mbox{ and S}=0.7, \nonumber \\
&  \mbox{P2 has I}=0.6 \mbox{ and S}=0.4, \nonumber \\ 
&  \mbox{P3 has I}=1.0 \mbox{ and S}=0.0. \nonumber 
\end{flalign}
 The probability of infection is computed as
\begin{flalign}
&pI_n=[  (1.0)(36) + (0.6)(30) + (0.3)(25) ]/216 = 0.285,\nonumber
\end{flalign}
\noindent This number $pI_n = 0.285$ approaches $p_2 = 1 - (5/6)^2 = 0.306$ since there is a fully infected participant and two other partially infected participants.  The updated partial infections for all seven participants in readiness for subsequent encounters are:
\begin{flalign}
&  \mbox{P1 has I}=  (0.7)(0.285) + 0.3 = 0.5 \mbox{ and S}= 0.5,  \nonumber \\
&  \mbox{P2 has I}= (0.4)(0.285) + 0.6 = 0.714 \mbox{ and S}= 0.286,  \nonumber \\
&  \mbox{P3 has I}= (0.0)(0.285) + 1.0 = 1.0 \mbox{ and S}= 0.0,  \nonumber \\
&  \mbox{P4-7 have I}= (1.0)(0.285) + 0.0 = 0.285 \mbox{ and S}= 0.715.  \nonumber
\end{flalign}

\subsection{Standard probability model}
\label{sec:standardpm}

\begin{figure}
\begin{centering}
\includegraphics[width=12.0cm]{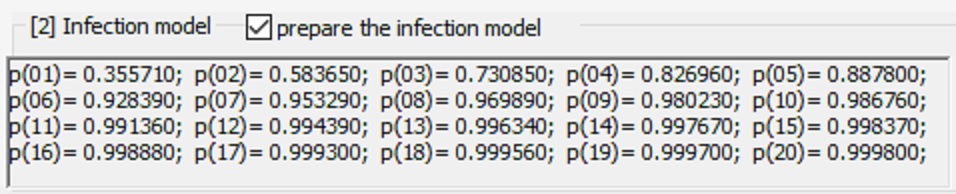}\\
\caption{Probabilities $p_n$ for various $n$ generated with $q=40$ trials by a Monte Carlo process.} \label{fig:key7}
\end{centering}
\end{figure}

The infection probability indeed follows a similar simple `probability count' philosophy as in the previous section but as obtained with a Monte Carlo process that generates the probabilities.  Each evaluation involves $q$ trials for one fully susceptible $S$ and twenty fully infected $I$ persons and it counts the times when the susceptible is co-located with any infected. For example, it does this on 40 separate occasions, i.e., once these forty trials are computed for the one susceptible and the twenty infected with results as in Figure~\ref{fig:key7}. The procedure checks to see whether a susceptible and one or more infected shared an encounter from those forty, ($q=40$ sampling). It is however fashioned in a more elaborate way to try to account for differing times spent in different sub-areas.

The random sampling is made to fit into 7200 seconds of time (two hours) but the random number indicating the location is chosen to weigh certain locations more.  It crudely captures that some areas of the store or establishment are more popular than others. For example, browsing a magazine rather than simply walking through some area of the store.  Random numbers are clustered to represent three types of areas: one where the person spends 2 seconds or another where the person spends 30 seconds, and yet another where the person spends 102 seconds in the selected area. The procedure is outlined in Appendix~\ref{appendix1}.

The constants in the figure are calculated prior to the run and give the probability of susceptible getting infected depending on how many people are infected in the store.  All visits are of the same duration so no attempt is made to model probability of infection through time with a Poisson process. This model of infection transmission is once again a very simple one.  When $n>20$ the probability of infection is assumed to be equal to one.  Note that a lower or higher rate of infection can be obtained by changing the sampling from 40 to a smaller or larger number. 

At any meeting place and time, three types of person can participate in the encounter: $I$ or infected; $S$ or susceptible and $R$ or immune (recovered). No action is taken if fewer than two people participate $n_p<2$. No action is taken if all are immune.  No action is taken if all are infected.  No action is taken if all are susceptible. No action is taken if there are no infected.  No action is taken if there are no susceptible.  Otherwise the aforementioned infection probability is selected for the number of infected and that real number is multiplied by the number of susceptible and then truncated to obtain the integer number that will become infected.  In no particular order that many susceptible are labelled as infected.

\section{Solution method}

The method adopted to optimize the allocation of requested visits is a Genetic Programming (GP) scheme developed by this author to discover a set of precise numerical constants that serve as coefficients of a polynomial representing the direct solution of the one dimensional homogeneous convection diffusion equation~\cite{howard:2001:gecco}.  This and other puiblications~\cite{Howard:2011convdiff}~\cite{Baber:2009:MSG} demonstrate that standard Genetic Programming trees are capable of computing very precise real numbers when and if needed.  When GP trees are evaluated they deliver a vector of real numbers of self-determined variable-length.

Table~\ref{tab:GP} lists the functions and terminals that comprise the GP tree. They manipulate two pointers $P_R$ and $P_Z$ to the output variable length vector of numbers. They also make use of two working memories, two real numbers,  $m_1$ and $m_2$.  Terminals are small and large numbers that become arithmetically manipulated.  Certain functions write to the variable length result vector, shrink it or expand it as the GP tree evaluates.  
\renewcommand{\baselinestretch}{1.4} 
\begin{table}[t!]
\begin{center}
\begin{tabular}[scale=0.55]{ | l | l ||  l | l | }
\hline
GP function name   				& description  					& GP function name    	& description   	\\ \hline
Constant(1)	 				& values (-127,-126,...,127,128)		& sConstant(1) 		& values (0,1,...,255)/255\\ \hline
AddNumber(2) 				& = L + R						& SubtractNumber(2) 	&  = L - R		\\ \cline{2-2}\cline{4-4}
MultiplyNumber(2) 				&  = L *  R						& DivideNumber(2)		& if ($\lvert$ R $\rvert<$ 1.0e-9) R=1\\\cline{2-2}
AverageNumber(2)			 	& = (L + R) / 2 					&  			 	& =  L / R			\\\hline
SubRecord(2) 				& = L							&ZeroRecord(2) 		& = L and increment $P_R$ \\
						&  if ($P_R  > 1$) decrease $P_R$		&				& if ($P_R  > P_{max}$) $P_R = P_{max}$\\
						& (there must be at least one element)		&				& $r_{P_R} = $ 0.00001\\\cline{2-2}\cline{4-4}
WriteRecord(2) 				& = R and increment $P_R$			& AddRecord(2) 		& = L and increment $P_Z$\\
						&  if ($P_R  > P_{max}$) $P_R = P_{max}$	&				& if ($P_Z  > P_{max}$) $P_Z = P_{max}$\\
						&$r_{P_R}$ = L					&				&$P_R = P_Z$ and $r_{P_R}$ = L\\\hline
GetMem1(2) 					& = $m_1$						& SetMem1(2)		& = L and $m_1$ = R\\\cline{2-2}\cline{4-4}
GetMem2(2) 					& = $m_2$						& SetMem2(2) 		& = R and $m_2$ = ($m_2$ +  L)/2\\
\hline
\end{tabular}
\end{center}
\caption{Constituent functions or building blocks of GP trees in~\cite{howard:2001:gecco} are of four varieties: number; arithmetic; record; and memory with number of arguments as shown in brackets: if (2), one is L or`left' input and the other R or`right' input (corresponding to left and right branches of the subtrees below the node) and (1) is a tree terminal or leaf. The resultant variable length vector of constants $r$ has elements $r_j$ that must never exceed $j= P_{max}$. Two pointers are manipulated $P_R$ and $P_Z$ current position and vector length respectively. Working memory locations are two: $m_1$ and $m_2$. In the above `increment' means to increase by one, and `decrease' to decrease by one. The symbol `=' denotes what the function returns to its parent node in the tree.}
\label{tab:GP}
\end{table}
The result of evaluating a GP tree is a variable length vector of real numbers.  These numbers can be of any size and can be positive or negative.  A subroutine then operates on these numbers to bound them as positive real numbers in size between 0.0 to 1.0 ~\cite{howard:2001:gecco}. For example, if a number in the vector is -21.27625 it now becomes 0.27625.  Also if a number is 29.00000 it now becomes 0.0001.

How is this vector of real numbers used? Consider 1704 visitation requests of Table~\ref{tab:data2}.  The real numbers vector is probably much smaller.  This is consulted from left to right as when a child reads words letter by letter.  Consider the outing request by a person is AD2, e.g. person 20 Monday in Figure~\ref{fig:datafile} wishing to go to Doctor's Surgery 2 at any time of the day.  The day is divided, into  eight two hour slots of time.  As each number ranges from 0.0 to 1.0 consider that this might be  0.2763.  This number would indicate prescription of the third slot of time since $2/8 < 0.2763 < 3/8$.  The allocation for that visit complete,  the next real number in the variable length vector gets consulted to deal with the next visit request.

As the variable length vector of numbers is usually shorter than the total number of visits requested, when the last element of the vector is reached it cycles back to the first number in the vector and continues until allocations for all 1704 visitations are dealt with, allocations of time as in Figure~\ref{fig:datafile}.  Excellent  solutions are achieved with small to moderate sized vectors.

Practical use of the method may require a small number of additional strategies. On some problems for a small number of initial generations the Darwinian fitness is set as the vector size. It stimulates production of large output vectors. When a desired variable length vector output size is reached by all members of the population the fitness is reset. From then onwards the fitness is the problem's Darwinian fitness, typically a measure of solution error.  The procedure is often not necessary but can become useful when required solution complexity or dimension is very high.  Once seeded in this way the solution vector will grow or shrink as crossover and mutation operations on the GP tree create new and improved GP trees.

As the method makes use of GP trees and standard GP, then all that we know about standard Genetic Programming including modularization options such as ADFs~\cite{koza:gp2}, and subtree encapsulation~\cite{roberts:2001:EuroGP} are applicable.  As will be revealed in the numerical experiments, the approach works, discovers good solutions, and appears to be quick on a portable computer, the compiled c++ Visual Studio 2019 executable delivering solutions in seconds for the proof of concept experiments.

\subsection{Measures of success and error}

Regardless of the infection model used the procedure is similar.  There will exists a taxonomy of persons by some parameter(s), for example, age group. Information about the likely degree of infection for different taxonomy classes is input to the computations.  As time progresses and visits take place, people get infected.  At the end of the three days, account is taken of who is infected and their prior health level.  This calculates how many people in the proof of concept study will become infected a few weeks hence and how many may die.  There are many assumptions but if correctly taken they drive Genetic Programming to discover allocations that reduce hospitalizations and fatalities.  Central to Darwinian fitness are such calculations of who and how many will perish or fall very ill and use the Intensive Care Unit (ICU).

\subsubsection{Darwinian fitness for the partial infection probability model}

\begin{figure}
\begin{centering}
\includegraphics[width=10.0cm]{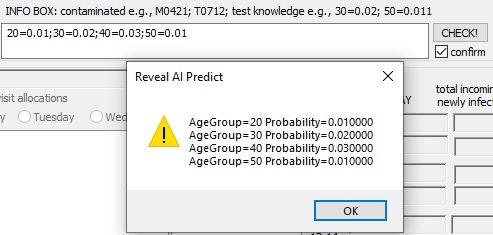}\\
\caption{Prior probability of infection by taxonomic group are entered for the runs. In this implementation taxonomy is assumed to depend on age group.} \label{probabinfect1}
\end{centering}
\end{figure}

COVID-19 testing is unlikely to become universal and frequent for all citizens but limited testing and other measurements are sufficient to serve as indications of the degree of infection likelihood for sectors, partitions, of the population.  The proof of concept uses age as taxonomy.  Figure~\ref{probabinfect1} reveals how this works. If testing reveals that more people of age groups A or B have COVID-19 than of age groups C or D, then percentages are entered at the start of numerical experiments. For example, 20=0.03; 30=0.01 means that the 20-year-olds have a suspected level of COVID-19 infection of three percent, while the 30-year-olds are suspected to have one percent.
Each simulation proceeds from Monday to Wednesday and for each day from morning to night (8 time slots). At each time slot all twelve establishments are considered and calculations of partial infection update the partial infection level of all.  At the end of the day, a calculation is made to determine who should go into self-isolation and participate no longer in the simulation. The rules to identify those persons are presented in the left side of Table~\ref{tab:IsoOut}.

At the very end of the simulation another procedure calculates the total number of hospitalized in the Intensive Care Unit (ICU), denoted by the symbol $N_H$.  These are those in hospital who eventually recover but who never the less put pressure on the health service. The procedure also calculates those who unfortunately pass away, denoted by the symbol $N_D$.  The rules to determine these two numbers are presented in the right side of Table~\ref{tab:IsoOut}. Note the assumption is made here that {\it both} the state of health and the age of the subject correlate similarly but are assumed to be separate causal factors.
\renewcommand{\baselinestretch}{1.4} 
\begin{table}[t!]
\begin{center}
 \begin{tabular}[scale=0.55]{ | c c c || cc c c c | }
\hline
 \multicolumn{3}{|c||}{self-isolation conditions} & \multicolumn{5}{c|}{outcome rules}\\
\hline
age 	& infected 	& health 	&age 			& infected 		& health 	& actions 	& outcomes 	\\ \hline 
 20  	& $I >$ 0.97	&  		& 20  			& $I >$ 0.95		& 7.0-10.0 	& none 	& immune\\
           & 0.95-0.97	& 0.0-7.0	&			& 			& 3.0-7.0	&  ICU		& recovered\\ 
	&		&		&			&			& 0.0-3.0	&  ICU 	& death\\ \hline
 30  	& $I >$ 0.95	&  		& 30 			& $I >$ 0.9   	& 8.0-10.0	& none 	& immune\\
	&  0.92-0.95	& 0.0-7.0	&			& 			& 4.0-8.0	& ICU		&recovered\\
	&		&		&			&			& 0.0-4.0	&ICU		&death\\ \hline			
  40  	& $I >$ 0.92	&  		&  40			& $I >$ 0.85 	& 8.0-10.0	& none 	&immune\\
           &  0.87-0.92	& 0.0-7.0	&			&			& 4.0-8.0	&  ICU		&recovered\\
	&		&		& 			&			& 0.0-4.0	&  ICU		&death\\\hline
   50  	& $I >$ 0.85	&  		&  50			& $I >$ 0.8   	& 8.0-10.0	& none 	&immune\\
          	&  0.80-0.85	& 0.0-7.0	&			&			& 4.0-8.0	&  ICU 	&recovered\\
	&		&		&			&			& 0.0-4.0	&  ICU		&death\\\hline
    60  	& $I >$ 0.75	&  		& 60			& $I >$ 0.75 	& 9.0-10.0	& none 	& immune\\
         	&  0.70-0.75	& 0.0-7.0	&			&			& 5.0-9.0	& ICU 		&recovered\\
	&		&		&			&			& 0.0-5.0	&ICU		&death\\ \hline		 
    70 	& $I >$ 0.65	&  		& 70			& $I >$ 0.7  		 & 9.5-10.0	& none 	& immune\\
          	&  0.60-0.65	& 0.0-7.0	&			&			& 7.5-9.5	& ICU 		&recovered\\ 
	&		&		&			&			& 0.0-7.5	& ICU		&death\\ \hline
   80  	& $I >$ 0.65	&  		& 80			& $I >$ 0.65 	& 8.5-10.0	&  ICU 	&recovered\\
          	&  0.60-0.65	& 0.0-7.0	&			&			& 0.0-8.5	&  ICU		&death\\ \hline
\end{tabular}
\end{center}
\caption{Partial Probability Model: (left) Conditions consulted after each day to determine if the person will self-isolate and no longer participate in assigned visits. (right) The outcome rules are consulted at the end of the simulation to determine the number of deaths, $N_D$, and the number of hospitalized, $N_H$ (all hospitalized minus those who die); both are used to compute the Darwinian fitness.}
\label{tab:IsoOut}
\end{table}

The GP Darwinian fitness function $F_F$ is the measure of solution goodness. It combines $N_H$ and $N_D$ weighted by some desirability constant $W_C$.  $F_F$ is by choice a negative quantity because the evolution is set to maximize or make bigger this quantity, with a perfect score being zero implying  $N_H = N_D = 0$:
\begin{align}
        F_F =  -1.0[(1 - W_C) N_H + W_C N_D]  
\label{eq:fitness}
\end{align}
All of the numerical results in this proof of concept use $W_C = 0.65$, reflecting desire to reduce fatalities.  

\begin{figure}
\begin{centering}
\includegraphics[width=16.0cm]{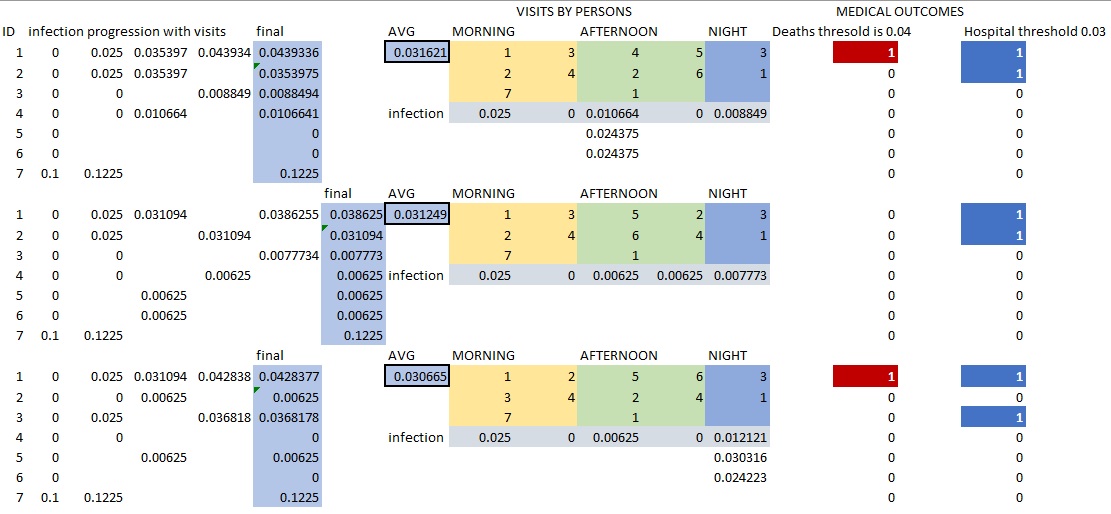}\\
\caption{In an example involving seven people changes in the allocation of visits for persons can result in higher death rates or hospitalizations with a lower final average infection.  If wanting to use partial infections as drivers it might be an idea to fuse various error norms: L$_2$, L$_1$, L${_\infty}$, etc.} \label{workedout1}
\end{centering}
\end{figure}

An alternative fitness function (not explored in this proof of concept) might compute the final average rate of infection: e.g., if $n_{2030}$ is the final average probability of infection of 20 and 30-year-olds and $n_{4060}$ the final average probability of infection of 40, 50 and 60-year-olds with  $n_{7080}$ being the final average probability of infection of 70 and 80-year-olds then a putative fitness measure is:
\begin{displaymath}
        F_F =  -1.0(W_1 n_{2030} + W_2  n_{4060} + W_3  n_{7080})  \nonumber
\end{displaymath}
\begin{displaymath}
W_1>0; W_2>0;W_3>0 \mbox{ and } W_1 + W_2 +W_3 = 1 \nonumber
\end{displaymath}
\noindent where $W_3>W_2>W_1>0$ reflects that the mortality of the young from COVI-D19 is low and that of the elderly is very high.  However, it suffices to show by the small worked out example of Figure~\ref{workedout1} (and as seen in the figures) that lower average partial infection does not necessarily reflect in smaller values of $N_H$ and $N_D$.
\subsubsection{Darwinian fitness for the standard probability model}

In this type of model persons can only become fully infected, $I$. Hence, the rules to determine $N_H$ and $N_D$ are different and only apply to those infected.  Additionally, the decision to self-isolate depends on the number of days that the person is infected and their health level.  Consider that some of the people in the input data file are already fully infected. 

Table~\ref{tab:IsoOutSPM} shows the decision schema for self-isolating persons as well as for deciding on $N_H$ and $N_D$ after the simulation is complete. Note the assumption is again made here that {\it both} the state of health and the age of the subject correlate similarly but are assumed to be separate causal factors.  The fitness measure for these computations is also equation~\ref{eq:fitness}.

\renewcommand{\baselinestretch}{1.4} 
\begin{table}[t!]
\begin{center}
 \begin{tabular}[scale=0.55]{ | c c c || c c c c | }
\hline
 \multicolumn{7}{|c|}{only for persons who become infected}\\
\hline
 \multicolumn{3}{|c||}{self-isolation conditions} & \multicolumn{4}{c|}{outcome rules}\\
\hline
age 	&days infected	& health 	&age 	& health 		&actions 	&outcomes \\ \hline 
20  	& $d_{Inf} > 1$	&$h < 5.0$  	& 20  	&$h > 7.0$		&none 	&immune\\
           & $d_{Inf} > 2$	&$h < 5.5$ 	&	&$3.0 <  h \le 7.0$	&ICU		&recovered\\ 
	&			&		&	&$h \le 3.0$		&ICU 		&death\\ \hline
30  	&$d_{Inf} > 1$	&$h < 6.0$ 	& 30 	&$h > 8.0$    	& none 	&immune\\
	&$d_{Inf} > 2$	&$h < 6.5$ 	&	&$3.5 <  h \le 8.0$ & ICU		&recovered\\
	&			&		&	&$h \le 3.5$		&ICU		&death\\ \hline		
40  	&$d_{Inf} > 1$	&$h < 6.5$ 	&  40	&$h > 8.0$ 		& none 	&immune\\
           &$d_{Inf} > 2$	&$h < 7.0$ 	&	&$4.0 <  h \le 8.0$	&ICU		&recovered\\
	&			&		& 	&$h \le 4.0$		&ICU		&death\\\hline
50  	&$d_{Inf} > 1$	&$h < 7.0$ 	&  50	&$h > 8.0$   	&none 	&immune\\
          	&$d_{Inf} > 2$	&$h < 8.0$ 	&	&$4.0 <  h \le 8.0$	&ICU 		&recovered\\
	&			&		&	&$h \le 4.0$		&ICU		&death\\\hline
60  	&$d_{Inf} > 1$	&$h < 7.0$ 	& 60	&$h > 8.5$   	&none 	& immune\\
         	&$d_{Inf} > 2$	&$h < 8.0$ 	&	&$4.5<  h \le 8.5$	&ICU 		&recovered\\
	&			&		&	&$h \le 4.5$		&ICU		&death\\ \hline	 
70 	&$d_{Inf} > 1$	&$h < 7.0$ 	& 70	&$h > 9.5$    	&none 	& immune\\
          	&$d_{Inf} > 2$	&$h < 8.0$ 	&	&$7.0 <  h \le 9.5$	&ICU 		&recovered\\ 
	&			&		&	&$h \le 7.0$		&ICU		&death\\ \hline
80  	&$d_{Inf} > 1$	&$h < 7.0$ 	& 80	&$h > 8.5$   	&ICU 		&recovered\\
          	&$d_{Inf} > 2$	&$h < 8.0$ 	&	&$h \le 8.5$		&ICU		&death\\ \hline
\end{tabular}
\end{center}
\caption{Standard Probability Model: (left) Conditions consulted after each day to determine if the person will self-isolate and no longer participate in assigned visits. (right) The outcome rules are consulted at the end of the simulation to determine the number of deaths, $N_D$, and the number of hospitalized, $N_H$ (all hospitalized minus those who die).}
\label{tab:IsoOutSPM}
\end{table}

\section{Results}

The numerical experiments compare the solution to three round robin uninformed allocations.  The first of these $comp1$ sends all those: (a)  requesting `morning' visits  or `any time' visits to the first morning time slot,  the  8:00 hrs to 10:00 hrs slot; (b) requesting `afternoon' visits to the first afternoon slot, the 12:00-14:00 hrs time slot; finally (c) requesting `night' visits are sent to the first night time slot, the 18:00 hrs to 20:00 hrs slot.  

The second of these uninformed allocations for comparison to results is a round robin of two, $comp2$. For case (a) it sends half the requests to the first slot of 8:00 hrs to 10:00 hrs and the other half to the 10:00 hrs to 12:00 hrs slot. For case (b) it sends half to the first time slot, 12:00 hrs to 14:00 hours, and the other half to the last slot of the afternoon, 16:00 hrs to 18:00 hours. For case (c) $comp2$ again sends half the requests to the first night slot, 18:00 hrs to 20:00 hrs, and the other half to the last night time slot, the 22:00 hours to 24:00 hours slot.

There is also a third uninformed allocation by round robin for comparison, $comp3$.  It is a round robin of three.  However, as the morning has only two time slots this sends two thirds of the requests for case (a) to the first slot and the other half to the second slot.  For cases (b) and (c) it uses all three time slots distributing the visitation requests evenly.  Comparison could be made to other allocations but these are broadly representative of what would transpire in the real world without the benefit of the solution and in conditions of partial lockdown or no lockdown.  It turns out, as expected, that $comp1$ results in the highest number of hospitalizations $N_H$ and deaths $N_D$ for the model problem and $comp3$ in the lowest as revealed in comparative experiments presented in this section.

Each experiment follows common GP practice executing a large number of parallel independent runs (PIRs). Each PIR differs in its  initial random seed (uses PC clock timer) seeding the population randomly and differently for each PIR.  PIRs can have different population sizes of GP trees.  All runs use 80 percent crossover and 20 percent mutation to generate new GP trees. This is a steady-state GP with tournament selection of four individuals to select a strong mate for crossover, whereby two GP trees swap branches, or the winner of the tournament simply mutates, with a kill tournament of two to choose the weaker GP tree to replace.   The maximum possible tree size for GP is set at 2000 and if this is exceeded then the shorter side of the crossover swap is taken. The maximum variable length vector size is set at 10,000 but never remotely approached: excellent solutions have vector sizes of between 10 and 100. It is an `elitist' GP for it does not destroy the fittest solution, i.e., solutions do not have a lifespan inside a PIR. All PIRs implement standard `vanilla' GP  with no parsimony pressure or any other non-standard approach.  This proof of concept study does not employ explicit reuse: ADF~\cite{koza:gp2} or subtree encapsulation~\cite{roberts:2001:EuroGP}. 

When a better solution emerges during the execution of a PIR, this is separately stored.  A PIR typically produces up to around ten of which two are highly fit and interesting.  As subsequent PIRs produce more solutions, all become ranked by fitness, a global list of solutions, from which the user can select one to inspect. For each solution, full details of the participants to all visitations, infection levels, numbers self-isolating, in ICU and sadly passed away can be inspected, as well as the variable length vector of real numbers that is the solution and visitation schedule and identity, age and level of infection of all participants recovered, in ICU or deceased. For the partial infection experiments it also computes partial infections by groups: young, middle aged and elderly every four hours. The comprehensive result panes in figures in this section merit closer inspection.  As PIRs build, a Pareto front picture can emerge of non-dominated solutions involving the two criteria $N_D$ and $N_H$.
\begin{figure}
\begin{centering}
\includegraphics[width=12.0cm]{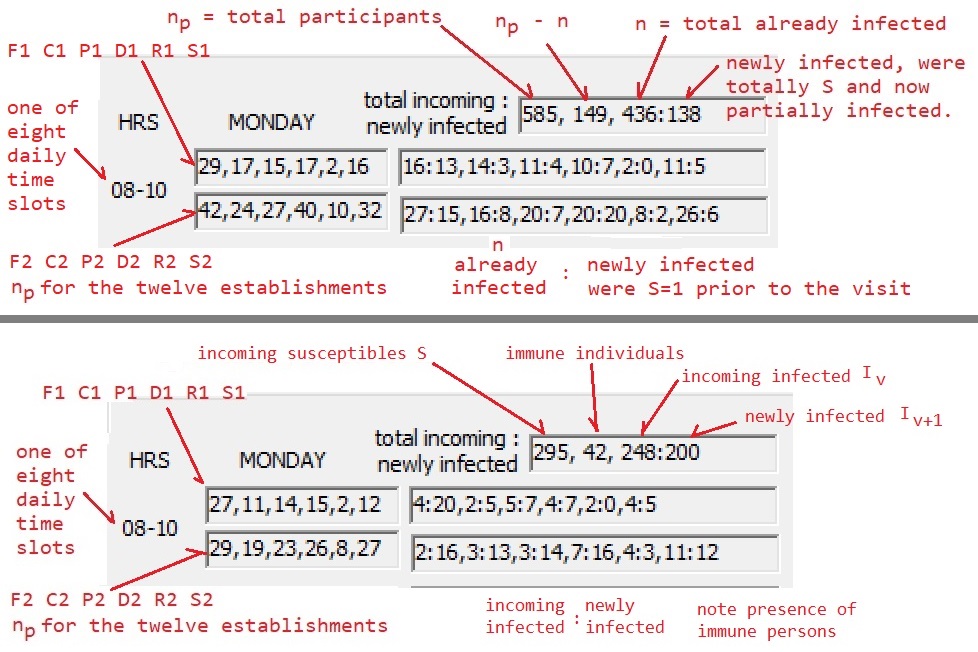}\\
\caption{The interface has solution details on number of: participants; infected; newly infected; visiting each establishment at each time slot and daily totals. Top: interpreting the output for experiments that use the partial infection model. Bottom: interpreting the output for experiments that use the standard probability model} \label{key1t}
\end{centering}
\end{figure}
\subsection{Numerical experiments with partial infection}

Table~\ref{tab:pires} shows the vast superiority of the solution to the uninformed round robin allocation schemes.  The superiority is marked and so there is no need to check this against other possible random allocation schemes.  It makes common sense that the Genetic Programming solution is always superior.

Note that with $s=6$ the problem becomes easier than with $s=4$ because the chances for meetings between people are lower.  This can be seen to be true for both round robin allocations and Genetic Programming solutions.  It also appears to be that the superiority of Genetic Programming solutions over round robin allocations is correlated with higher $s$. This is probably because the Genetic Programming has more degrees of freedom to discover a better solution.  Therefore, even if the intuitive idea is that with a smaller density of people there should be less need for the solution, this is not always the case. Note that in the cases where $s=6$, Genetic Programming managed to discover an allocation where $N_D=0$, with no deaths.  Perhaps such a counter-intuitive conclusion can be discerned that the solution is needed more so when the problem seems simpler.

Close inspection of the figures referred to in Table~\ref{tab:pires} reveals that the system of PIRs delivers a number of Pareto non-dominated solutions to choose from.  Also, there are solutions that achieve an identical result in terms of $N_D$ and $N_H$ but which are quite different.  Then one is also able to inspect the age and health of the fatalities, again this may be a tertiary factor that could come into play when selecting among the many equivalent solutions.  Finally, it can also be discerned that the average partial infection level and its final levels as shown in the figures do not always correlate with lower error and higher Darwinian fitness.

\renewcommand{\baselinestretch}{1.4} 
\begin{table}[t!]
\begin{center}
 \begin{tabular}[scale=0.55]{| c | c || r | r | l || r | r | r | l | }
\hline
 \multicolumn{2}{|c||}{parameters} & \multicolumn{3}{c||}{$comp1$, $comp2$ or $comp3$} & \multicolumn{4}{c|}{fittest GP solutions}\\
\hline
$pI_n$ input				& s 		&  $N_H$ 	& $N_D$ 	& figures  in text				&  $-F_F$ 	&  $N_H$ 	& $N_D$ 	& figures in text	\\ \hline 
  					&4 		&  70		& 52  		&$comp1$: see~\cite{inha2020}		&7.70	 	&9		&7		& Figure \ref{fig:pI04}\\ 
 20 = 0.01; 40 = 0.03;		&		&  44		& 23  		&$comp3$: Figure \ref{fig:pI03}		&7.70	 	&9		&7		&see~\cite{inha2020}\\
 50 = 0.02				&		&  		& 		&						&8.15		&14		&5		&see~\cite{inha2020}\\ \cline{2-9}
					&6		&  7		& 8  		&$comp3$: Figure \ref{fig:pI07}		&0.0	 	&0		&0		&Figure \ref{fig:pI08}\\ \hline
 					&4		& 72 		& 52  		&$comp1$: see~\cite{inha2020}		&20.05 	&35		&12		&Figure \ref{fig:pI12}\\
20 = 0.05; 30 = 0.05;		&		& 64 		& 41  		&$comp2$: see~\cite{inha2020}		&20.50	 &40		&10		&see~\cite{inha2020}\\
40 = 0.03; 50 = 0.05;		&		& 57 		&34   		&$comp3$: Figure \ref{fig:pI11}		&	 	&		&		& 			 \\ \cline{2-9}
60 = 0.01				&6		& 69 		&49   		&$comp1$: see~\cite{inha2020}		&1.05	 	&3		&0		&Figure \ref{fig:pI17}\\
					&		&52  		&29   		&$comp2$: see~\cite{inha2020} 	&1.65	 	&1		&2		&see~\cite{inha2020}\\
					&		&32  		&17  		&$comp3$: Figure \ref{fig:pI16}		&1.70	 	&3		&1		&see~\cite{inha2020}\\ \hline
\end{tabular}
\end{center}
\caption{Summary of results attained by the numerical experiments that made use of the partial infection model. With $s=4$ the solution is three times better than a round robin allocation and with $s=6$ it is ten times better.}
\label{tab:pires}
\end{table}

\subsection{Numerical experiments with full infection}

A different set of computations to the real world problem is included here for completion.  It pertains to knowing who is infected {\em a priori} and trying to optimize what could have happened had their visits been scheduled differently. It could either be useful to a retrospective study or, alternatively, such computations can potentially address the same real world problem but only by the carrying out of a plethora of experiments with different seeds of infected individuals, assumed infected according to some taxonomy knowledge of level of infection in the population, and then averaging the results in some way to prescribe safer allocations of visits, as alternative to the experiments (presented in the previous section) with the partial infection model.

\begin{figure}
\begin{centering}
\includegraphics[width=14.0cm]{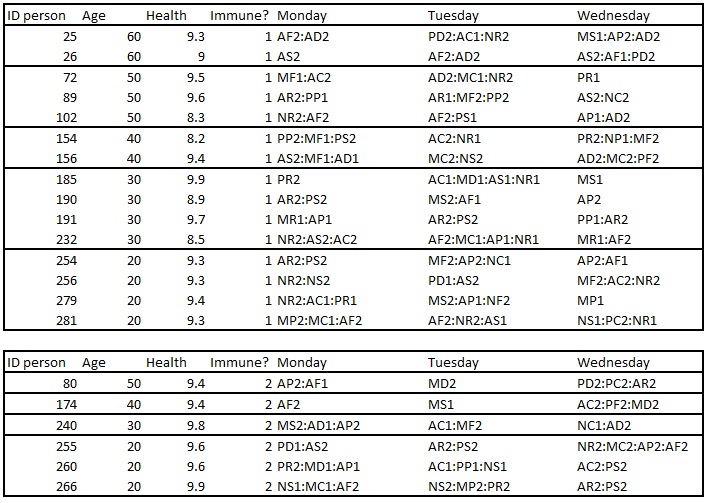}\\
\caption{Part of the dataset: (top) those who come infected {\em a priori} driving the `full infection' numerical experiments and (bottom) a few with immunity or recovered who cannot get infected. } \label{fig:apriori}
\end{centering}
\end{figure}

A set of 15 persons, as shown in Figure~\ref{fig:apriori}, out of the 282 in the proof of concept data, as described in section~\ref{section:data}, come already fully infected {\em a priori} to drive the computations\footnote{Note that in the partial infection model experiments, the field `Immunity?' is completely ignored but not here, with values: 0 for susceptible; 1 for already infected and 2 for already immune or recovered.}.    This is about 5.3 \% of total people, and they undertake 105 visits or 6.2 \% of the 1704 total visits that appear in the proof of concept dataset. Individuals with good health levels are chosen as already infected.  For completion, the figure also shows six people who have immunity and therefore cannot become infected in the computations. They are included in computations but there is no need for them.  

\begin{figure}
\begin{centering}
\includegraphics[width=12.0cm]{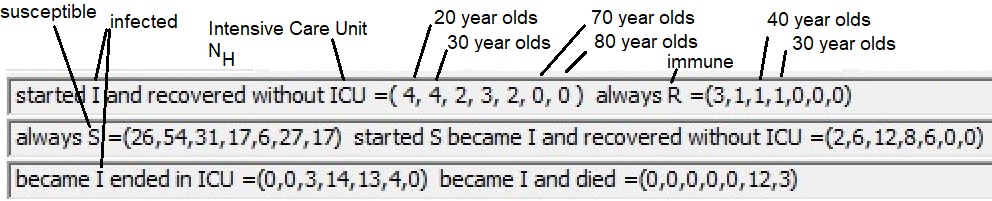}\\
\caption{Output that facilitates comprehension of six classes of starting and outcome combinations.} \label{fig:key8}
\end{centering}
\end{figure}

It is envisaged that the optimization problem by GP is challenging.  Hence, the infection model is implemented at different values of $q$ (see section~\ref{sec:standardpm} and appendix~\ref{appendix1}) to allow the solution room to make gains but also to understand the dynamics under various levels of contagion (perhaps representing adherence to social distancing and use of masks).  There are eight types of combinations of the events that can befall a person:
\begin{enumerate}
\item person comes to the simulation already infected and does not use the Intensive Care Unit (ICU);
\item person comes to the simulation already infected and does make use of the ICU (adds to $N_H$);
\item person comes to the simulation already infected passes through the ICU or not and dies (adds to $N_D$);
\item person comes to the simulation immune (has had the disease before and has recovered) and stays that way;
\item person comes to the simulation susceptible and does not get infected;
\item person comes to the simulation susceptible, gets infected but does not use the Intensive Care Unit (ICU);
\item person comes to the simulation susceptible, gets infected and does make use of the ICU (adds to $N_H$);
\item person comes to the simulation susceptible, gets infected, goes or not to ICU and dies (adds to $N_D$);
\end{enumerate}
Figure~\ref{fig:key8} illustrates how to interpret that part of the result figures pointed to in Table~\ref{tab:fullres}.  As there are generally no cases of infected ending in ICU or dying only six result summaries show as in the figure.

The results of all numerical experiments that use the full infection model are as in Table~\ref{tab:fullres}.  The difficulty with this infection model is that it multiplies the probability of infection by the number of susceptible and then it simply truncates the result to the lower integer. Moreover, it simply goes down the list of susceptible infecting the first bunch it sees up to that integer number.  This gives the solution opportunities to play games. For example if $q=30$ then $p(9)=0.947$ and if there is only one susceptible and nine infected then the susceptible will not become infected.

Notwithstanding the weaknesses of such computations and their erroneous assumptions this can be said about the Table~\ref{tab:fullres} figures.  As $q$ is increased then the GP search becomes much harder. Best solutions come from bigger GP populations run for longer (longer numbers of generations).  Also the advantage of the solution over the round robin assignments is less valuable than at lower $q$ values.  This is to be expected because at high $q$ the disease is far more contagious and there is not much room for the optimization.  Note that with $q=30$ all round robin solutions have the same $N_H$ and $N_D$ possibly indicating little room for schedule optimization by the solution.
\renewcommand{\baselinestretch}{1.4} 
\begin{table}[t!]
\begin{center}
 \begin{tabular}[scale=0.55]{|  c || r | r | l || r | r | r | l | }
\hline
 \multicolumn{1}{|c||}{parameters} & \multicolumn{3}{c||}{$comp1$, $comp2$ or $comp3$} & \multicolumn{4}{c|}{fittest GP solutions}\\
\hline
 $q$ (Monte Carlo) 		&  $N_H$ 	& $N_D$ 	& figures  in text				&  $-F_F$ 	&  $N_H$ 	& $N_D$ 	& figures in text\\ \hline 
 5				& 70		& 53		&$comp1$: see~\cite{inha2020}		&1.05		&3		&0		&Figure \ref{fig:full15}\\
 				& 49		& 26		&$comp2$: see~\cite{inha2020}		&1.35		&2		&1		&Figure \ref{fig:full16}\\
 				& 31		& 17		&$comp3$: Figure \ref{fig:full14}	&		&		&		&\\\hline
10 				&  71		& 53 		&$comp1$: see~\cite{inha2020}		&21.00	&34		&14		& Figure \ref{fig:full04}\\
				&  69		& 52  		&$comp2$: see~\cite{inha2020}		&21.30	&33		&15		& see~\cite{inha2020}\\
				&  64		& 40		&$comp3$: Figure \ref{fig:full03}	&21.65	&34		&15		& see~\cite{inha2020}\\ \hline
30				&  71		& 53  		&$comp1$: see~\cite{inha2020}		&44.60	&68		&32		& Figure \ref{fig:full10} \\ 
				&  71		& 53  		&$comp2$: see~\cite{inha2020}		&44.90	 &67		&33		& see~\cite{inha2020}  \\ 
				&  71		& 53  		&$comp3$: Figure \ref{fig:full09}	&	 	&		&		& \\ \hline
\end{tabular}
\end{center}
\caption{Summary of results attained by the numerical experiments that made use of the full infection model. With $q=4$ ( see Appendix~\ref{appendix1} and the text), the solution is between two and three times better than simple round robin allocations.}
\label{tab:fullres}
\end{table}
\section{Conclusions}
\label{sec:conclusions}

Large improvements in both lower mortality and lower use of the ICU are available with the solution for the proof of concept data described in section~\ref{section:data} in computations with both infection models.  The improvements as compared to round robin allocations in tables: Table~\ref{tab:pires} and Table~\ref{tab:fullres} clearly show it.  The parameters $s$ and $q$ respectively in the two numerical experiments have a similar but inverse influence as the former gives the number of possible sub-locations that can be occupied while the latter is the number of checks performed inside the Monte Carlo calculation that detects co-presence.  However, even at low $s$ and at high $q$, the worst possibilities, the solution keeps outperforming round robin by a significant margin.  At high $s$ and low $q$ the solution really shines and outperforms round robin by a very considerable margin.

It augers well because with good cleaning at locations, washing of hands and even moderate social distancing, the rate of infection is expected to be low (corresponding to higher $s$ and lower $q$ in a sense).  Of course, a more serious real-world model needs to be considered in further research.  Such a model would need to consider:
\begin{enumerate}
\item is it correct that a difference in infection rates exists between taxa? if so, what is this taxonomy and how to construct it? is it possible to discover this taxonomy through COVID-19 testing and other data collection?
\item is it possible to develop a reasonably accurate infection model for certain stores and places that people wish to, or need to, frequent? 
\item the infection model must also include a dynamic related to object contamination and transmission through contact of surfaces and objects;
\item travel by public transport to such locations also needs to be accounted for by the model;
\item time at the location, estimated, should be taken into consideration perhaps by a Poisson distribution like infection and contamination function;
\item speed of motion of different types of people through a location of visit may be another characteristic affecting infection rates;
\item what is the maximum number of requests for outings by people that makes the solution very effective? what level of demand makes it less effective?
\item  is the round robin a fair reflection of how people wish to go out in the unrestricted normal case? are there invariant principles gathered from mobile phone roaming data that could inform how and when people go out?
\item  what is the effect of non-compliance on the solution? could the solution account for it and still be gainful?
\item  the solutions arrange into Pareto non-dominated sets involving $N_D$ and $N_H$: what is the difference between these and also between equivalent solutions, same score of $F_F$? is there a difference between such solutions of further interest?
\item the solution is designed to work even with very little idea or precision about the rates of infection and contamination as it aims for an improvement rather than precise values. How valid are these assumptions?
\item the solution out pefoms round robin but how does it fare in terms of infection rate against strict lockdowns, or exiting lockdown with phased lockdown strategies suggested in~\cite{delockdown}?
\item can economic, psychological and other benefits of the solution be quantified to understand the cost benefit analysis of adopting it versus strict lockdown policies?
\item will people be willing to undergo numerous lockdowns waiting for the availability of a vaccine if the solution is adopted?
\end{enumerate}

Genetic Programming is known to scale well with problem size.  However, even if millions of people were considered there would be a degree of clustering that could be treated differently by different discovered solutions.  The partial infection model is developed here for the first time.  If there is something similar in the literature then this author does not know of it.  It must be tested and developed better. It is possible that the pessimistic approach of multiplying the terms of worst case is not entirely reasonable.  Lastly, how well infections can be avoided will depend on the number of visitations, the number of establishments visited, the number of people and the frequency of visits.  In this research Genetic Programming was handed a tough challenge as the number of visitations was in the order of 2000 and the times slots very few per day. Also the number of establishments was only a few.

In general, it is said that washing one's hands is far more effective than social distancing.  It is probably true that contamination is more important than person to person transmission.  Contamination can be easily incorporated into the model.  Although the model is incipient, it is considered something few have considered if any.  Most research is involved in exploiting data sources to predict infection levels or explain the disease.  This contribution is different in nature because its deployment could generate data and would also need only tendency of disease data.  This contribution is markedly different from contact tracing approaches that are {\it reactive}.  The work described here and its implementation would be {\it proactive} but it could also inform and be informed by contact tracing.

\appendix
\newpage
\section{Pseudo-code for the Monte Carlo routine}
\label{appendix1}
The following pseudo-code with $q=40$ produces the numbers of Figure~\ref{fig:key7}:

\begin{tabular}[scale=0.55]{|l|}
\hline
do m, 20 times (iterate over 20 infected)\\
\hspace{1cm}nInfection[m] =0  (counts the encounters for this many infected)\\
end iteration index m\\
do 100,000 times (use a big number for good accuracy)  \\
\hspace{1cm}do k, q times (typicaly q is between 10 and 40) \\
\hspace{1.5cm}21 times (calculate for 1 susceptible and 20 infected) call index m \\
\hspace{2cm}i= random number; range (1:7200) \\
\hspace{2cm}if  i < 600 (300 2x2 m2 areas, spent 2 seconds in each: corridors traversed quickly)\\
\hspace{3cm}j = i \\
\hspace{3cm}trial[k]][m] = j/2 \\
\hspace{2cm}else if  i < 2100 (50 2x2 m2 areas, spent 30 seconds in each: shelves) \\
\hspace{3cm}j = i – 600 \\
\hspace{3cm}trial[k][m] = 300 + j/30\\
\hspace{2cm}else (50 2x2 m2 areas, spent 102 seconds in each: popular shelves) \\
\hspace{3cm}j = i – 2100 \\
\hspace{3cm}trial[k][m] = 350 + j/102 \\
\hspace{2cm}end if\\
\hspace{1cm}end iteration index k\\
\hspace{1cm}do k, q times (did infected $I$ meet susceptible $S$ in any k trial)\\
\hspace{2cm}do m, 20 times (check 20 infected)\\
\hspace{3cm}if trial[k][m] == trial[k][0] (0 is the susceptible individual)\\
\hspace{4cm}do k2, 20 times (check 20 infected)\\
\hspace{5cm}bMet[k2] = true (records meet between an $S$ and the $I$s)\\
\hspace{4cm}end iteration index k2\\
\hspace{3cm}end if\\
\hspace{2cm}end iteration index m\\
\hspace{1cm}end iteration index k\\
\hspace{1cm}do m, 20 times (iterate over 20 infected)\\
\hspace{2cm}if bMet[m] == true (was there {\it any} meeting?))\\
\hspace{3cm}nInfection[m] = nInfection[m] + 1  (adds to the global count)\\
\hspace{2cm}end if\\
\hspace{1cm}end iteration index m\\
end Monte Carlo iteration\\
do m, 20 times (iterate over 20 infected)\\
\hspace{1cm}dInfectProbability[m] = ((double)nInfection[m])/100000.0\\
end iteration index m\\
\hline
\end{tabular}

\begin{tabular}[scale=0.55]{|r| r| r| r|r |}
\hline
random number 	&interval	& number of areas	&visit time 	 	&size of area\\\hline
0-600 			&600		&300			&2 secs 		&2x2 m$^2$ squares\\
601-2100 		&1500		&50			&30 secs 		&2x2 m$^2$ squares\\
2101-7200		&5100		&50			&102 secs 		&2x2 m$^2$ squares\\\hline
7200			&7200		&400			&7200 secs		&1600 m$^2$ walk surface\\\hline
\end{tabular}
 
\section{Appendix: Simple infection models for the computations}

\subsection{When chance encounters underlie infection}

When a susceptible $S$ individual meets an infected individual $I$ there is a possibility of transmission of the disease\footnote{as contrasted with place contamination and contact with things, and the abating of infection by the washing of hands} from $I$ to $S$.  This is usually expressed as an infection probability $p_n$ where $n$ is the number of infected individuals present at the location of the encounters between the susceptible individual and $n$ infected individuals. Sophisticated relations can be determined in function of the number of encounters between a susceptible and one or more infected that also incorporate time of exposure modelled as a Poisson process \cite{bib:grahammedley}\footnote{Personal communication: `the risk of transmission increases non-linearly with the number of infected and with time. Suppose I go to the garden centre for one hour and there is one other person, and it gives me a rate, p, of being infected each hour we are there. So the probability of me being infected at the end of the hour is (1-exp(-p*1)), and if we are there 2 hours: (1-exp(-p*2)). If there are n people there then the time I spend contacting each person drops, so the risk per person drops. A reasonable assumption is that the risk with n people is $p*n^c$, where $c<1$. This gives you a contamination function.'}.  An important characteristic of such relations is that as the number of infected individuals $n$ grows then $p_n$ increases but not linearly, so that for example one can expect $p_2 < 2 p_1$. 

\begin{figure}
\begin{centering}
\includegraphics[width=12.0cm]{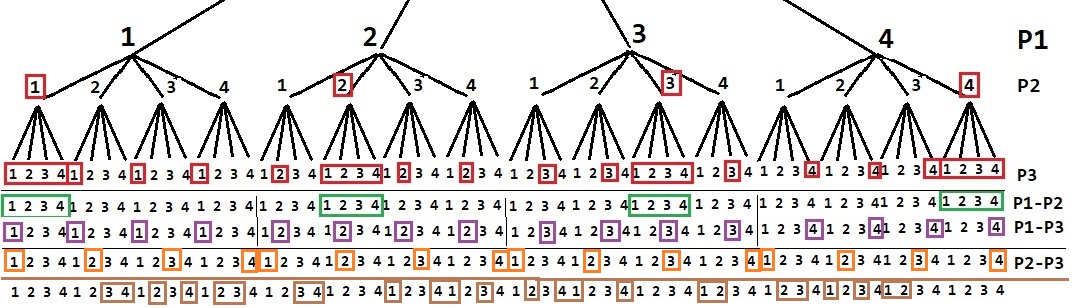}\\
\caption{Location probability tree involving three individuals P1, P2, and P3 who may occupy one of four possible locations: 1, 2, 3 or 4.} \label{figlocations1}
\end{centering}
\end{figure}

This proof of concept assumes a trivially simple infection function based on counting co-locations between $S$ and $I$. These increase with the number of infected but the ratio of these to the total possible encounters never exceeds one, that is, as $n \to\infty$ then $p_\infty \to 1.0$.  The essence of this behaviour can be crudely emulated with Monte Carlo or, as discussed here, by counting on simple probability trees of Figure~\ref{figlocations1}. We ignore dependency on time of exposure to assume all visits to places are of similar duration. The Genetic Programming approach that uses such infection functions, however, is general and able to incorporate any candidate infection function.

From this figure, first, consider the case of two individuals only: P1 and P2.  Individual P1 is susceptible while P2 is infected.  The number of encounters between these two, the red boxes at the P2 level in the figure, or where both individuals share a same location, is four and the total number of possible outcomes is sixteen.  Hence, the opportunity for an encounter taking place and therefore for infection, assuming each individual should have an equal presence for each of the four locations and spend the same amount of time at any visited, is $p_1 = 0.25$.  Next consider the case where three individuals participate with P1 susceptible and P2 and P3 both infected.  Now we recognize opportunities for all three individuals to exist at the same location, and also opportunities for P1 to share a location with either infected P2 or P3.  If we count such opportunities we arrive at $p_2 = 28/64 =0.4375$ and we verify that  $p_2 < 2 p_1$. For the particular scenario of Figure~\ref{figlocations1} the coarse emulation leads to a simple relation to obtain the probability of encounter for any $n$ that is:  $p_n = 1 - (3/4)^n$.  Considering rather large $n$  it tends to one:  $p_{20} = 1 - (3/4)^{20} = 0.9968$.

\subsection{Introducing: partially infected people}
\label{sec:partial}

We now introduce perhaps what may be an unusual idea. The real world objective of this initiative is to leverage off COVID-19 testing. Imagine that upon testing for COVID-19 the incidence is found to be higher in age groups 20 and 50. It is unlikely that COVID-19 testing will be undertaken by all people all of the time.  Moreover, tests are still unreliable.  Hence, we need to work with partial knowledge.  An option to explore might be to work directly with the probable levels of infection that are informed by the testing for different groups of people as organized in some taxonomy:  for example, consider the probability of infection to be 5 percent and 8 percent respectively and zero in other age groups?\footnote{without loss of generality, this work assumes that age group confers to persons variable risks of already having the infection}.  It motivates use of a `partial infection' but why?  If we knew who was or was not infected we would let them out or keep them in isolation.  However, as we cannot test every single person, it is useful to assign to them uncertainty, a probability level\footnote{alternatively, we could use fuzzy set membership}. In such a case, we must consider that any of the people in Figure~\ref{figlocations1} namely P1, P2 or P3 may be partially infected (and partially susceptible). 

Figure~\ref{figlocations1} shows the encounters between persons, i.e., two people sharing one location (1,2,3,4) at the same moment in time, thus coming into contact with each other.  We are especially interested in two encounters: P1-P2 and  P1-P3 as we can assume that one person P1 is susceptible while the other two are infected.  Note that under permutations assigning $I$ and $S$ there is no need to count encounters P2-P3 as these would represent two infected, and neither of interest are encounters between two susceptible.  The thirty two encounters that determined the probability of infection result in the union that gives twenty eight encounters in the figure.  Each of P1-P2 and P1-P3 result in 16 encounters but 4 encounters are shared. So only 28 out of 64 possible combinations of three locations are or interest.  This ratio gives a crude probability of infection $p_2 = 28/64 =  1 - (3/4)^{2} = 0.4375$. We now consider permutations of $I$ and $S$ components of partially infected persons in such basic probability tree count calculations as will become apparent in the below examples\footnote{Out of interest the number of non encounters (three different locations for the three persons) are twenty four: 123, 124, 132, 134, 142, 143, 213, 214, 231, 234, 241, 243, 312, 314, 321, 324, 341, 342, 412, 413, 421, 423, 431, 432.}. 

\subsubsection{An example involving three people}
As an example, consider the interactions between three people with partial infection:
\begin{flalign}
&  \mbox{P1 has I}=0.03 \mbox{ and S}=0.97, \nonumber \\
&  \mbox{P2 has I}=0.04 \mbox{ and S}=0.96, \nonumber \\
&  \mbox{P3 has I}=0.01 \mbox{ and S}=0.99. \nonumber
\end{flalign}
To arrive at an infection probability we need to consider contributions from the following six possibilities\footnote{there is no need to consider SSS or III}:  ISI, IIS, ISS, SII, SSI, SIS. Here are the numerical contributions from each term: 
\begin{flalign}
& \mbox{ISI}=[(0.96)(0.03)(16-4) + (0.96)(0.01)(16-4) + (0.96)(0.03)(4)]/64 = 0.0090, \nonumber \\
& \mbox{IIS}=[(0.99)(0.03)(16-4) + (0.99)(0.04)(16-4) +(0.99) (0.04)(4)]/64 = 0.0155,\nonumber \\
& \mbox{ISS}=[(0.99)(0.03)(16)]/64  = 0.0074,\mbox{ subsumed in ISI and with } 0.99 \mbox{ in IIS} \nonumber\\
& \mbox{SII}=[(0.97)(0.01)(16-4) + (0.97)(0.04)(16-4) + (0.97)(0.04)(4)]/64 = 0.0115, \nonumber\\
& \mbox{SSI}=[(0.97)(0.01)(16)]/64  = 0.0024, \mbox{ subsumed in SII and with } 0.96 \mbox{ in ISI} \nonumber\\
& \mbox{SIS}=[(0.99)(0.04)(16)]/64  = 0.0099, \mbox{ subsumed in IIS and with } 0.97 \mbox{ in SII} \nonumber
\end{flalign}
Note that the encounters of overlap where all three persons are in the same location give preference to the largest infection, hence the third term which multiplies by four, the overlapping cases (see Figures~\ref{figlocations1},~\ref{figlocations2} and ~\ref{figMASSIVE} ) The total probability of infection is the maximum of ISI, IIS and SII or 0.0155.  Now we compute the new partial infections for our three participants:
\begin{flalign}
&  \mbox{P1 has I}=(0.97)(0.0155) + 0.03 =0.0450 \mbox{ and S}=0.9550,  \nonumber \\
&  \mbox{P2 has I}=(0.96)(0.0155) + 0.04 =0.0549 \mbox{ and S}=0.9451,  \nonumber \\
&  \mbox{P3 has I}=(0.99)(0.0155) + 0.01 =0.0253 \mbox{ and S}=0.9747.  \nonumber
\end{flalign}
 
\subsubsection{A second example involving three people}
Here is a second example, consider the interactions between three people with partial infection:
\begin{flalign}
&  \mbox{P1 has I}=0.95 \mbox{ and S}=0.05, \nonumber \\
&  \mbox{P2 has I}=0.98 \mbox{ and S}=0.02, \nonumber \\
&  \mbox{P3 has I}=0.01 \mbox{ and S}=0.99. \nonumber
\end{flalign}
which approaches the situation of one susceptible person, P3, meeting two infected: P1 and P2.  Again to arrive at an infection probability we  consider contributions from the following six possibilities:  ISI, IIS, ISS, SII, SSI, SIS. Here are the numerical contributions from each term: 
\begin{flalign}
& \mbox{ISI}=[(0.02)(0.95)(16) + (0.02)(0.01)(12)]/64  = 0.0048, \nonumber \\
& \mbox{IIS}=[(0.99)(0.95)(12) + (0.99)(0.98)(16)]/64  = 0.4189,\nonumber \\
& \mbox{SII}=[(0.05)(0.01)(12) + (0.05)(0.98)(16)]/64  = 0.0115. \nonumber
\end{flalign}
The total probability of infection is the maximum of ISI, IIS and SII or 0.4189. This number is close to  $p_2 = 28/64 =0.4375$.  Now we compute the new partial infections for our three participants:
\begin{flalign}
&  \mbox{P1 has I}=(0.05)(0.4189) + 0.95 = 0.9709 \mbox{ and S}= 0.0291,  \nonumber \\
&  \mbox{P2 has I}=(0.02)(0.4189) + 0.98 = 0.9884 \mbox{ and S}= 0.0116,  \nonumber \\
&  \mbox{P3 has I}=(0.99)(0.4189) + 0.01 = 0.4247 \mbox{ and S}= 0.5753.  \nonumber
\end{flalign}
Note that in spite of P3 already carrying a small probability of infection, its new infection level $0.4247 < 0.4375$.  This is because others were not one hundred percent infected and the result was close to but less than  0.4375.

\subsubsection{A third example involving three people}
Here is a third example, consider the interactions between three people with partial infection:
\begin{flalign}
&  \mbox{P1 has I}=0.01 \mbox{ and S}=0.99, \nonumber \\
&  \mbox{P2 has I}=1.00 \mbox{ and S}=0.00, \nonumber \\
&  \mbox{P3 has I}=0.01 \mbox{ and S}=0.99. \nonumber
\end{flalign}
which approaches the situation of the interactiion of the non-infected with a fully infected person.  Again to arrive at an infection probability we  consider contributions from the following six possibilities:  ISI, IIS, ISS, SII, SSI, SIS. Here are the numerical contributions from each term: 
\begin{flalign}
& \mbox{ISI}=[(0.00)(0.01)(16) + (0.00)(0.01)(12)]/64 = 0.0, \nonumber \\
& \mbox{IIS}=[(0.99)(0.01)(12) + (0.99)(1.00)(16)]/64 = 0.2493,\nonumber \\
& \mbox{SII}=[(0.99)(1.00)(16) + (0.99)(0.01)(12)]/64 = 0.2493.\nonumber
\end{flalign}
The total probability of infection is the maximum of ISI, IIS and SII or 0.2493. This number is very close to  $p_1 = 4/16 =0.25$.  Now we compute the new partial infections for our three participants:
\begin{flalign}
&  \mbox{P1 has I}=(0.99)(0.2493) + 0.01 = 0.2568 \mbox{ and S}= 0.7432,  \nonumber \\
&  \mbox{P2 has I}=(0.00)(0.2493) + 1.00 = 1.0000 \mbox{ and S}= 0.0000,  \nonumber \\
&  \mbox{P3 has I}=(0.99)(0.2493) + 0.01 = 0.2568 \mbox{ and S}= 0.7432.  \nonumber
\end{flalign}
Notice that P1 and P3 have an infection level $ 0.2568 > 0.2500$ this is because the people were already one percent infected.

\subsubsection{A first example involving four people}

An example involving four people is illustrated with the help of Figure~\ref{figlocations2}, consider the interactions between three individuals with partial infection:
\begin{flalign}
&  \mbox{P1 has I}=0.01 \mbox{ and S}=0.99, \nonumber \\
&  \mbox{P2 has I}=0.98 \mbox{ and S}=0.02, \nonumber \\
&  \mbox{P3 has I}=0.03 \mbox{ and S}=0.97, \nonumber \\
&  \mbox{P4 has I}=0.05 \mbox{ and S}=0.95. \nonumber 
\end{flalign}
\noindent Again to arrive at an infection probability we  consider contributions from the following: SIII, ISII, IISI, IIIS. Here are the numerical contributions from each term: 
\begin{flalign}
& \mbox{SIII}=[(0.99)(0.98)(64) + (0.99)(0.03)(36) +  (0.99)(0.05)(48)]/256 = 0.2560,\nonumber \\
& \mbox{ISII}=[(0.02)(0.01)(36) + (0.02)(0.03)(48) +  (0.02)(0.05)(64)]/256 = 0.0004,\nonumber \\
& \mbox{IISI}=[(0.97)(0.01)(36) + (0.97)(0.98)(64) +  (0.97)(0.05)(48)]/256 = 0.2481,\nonumber\\
& \mbox{IIIS}=[(0.95)(0.01)(36) + (0.95)(0.98)(64) +  (0.95)(0.03)(48)]/256 = 0.2394.\nonumber
\end{flalign}
The total probability of infection is the maximum of SIII, ISII, IISI and IIIS, or 0.2560. This number is very close to  $p_1 = 1 - 3/4 = 1/4 =0.25$. It is a higher value because all participants contribute a significant level of infection.  Now we compute the new partial infections for our four participants:
\begin{flalign}
&  \mbox{P1 has I}=(0.99)(0.2560) + 0.01 = 0.2634 \mbox{ and S}= 0.7366,  \nonumber \\
&  \mbox{P2 has I}=(0.02)(0.2560) + 0.98 = 0.9851 \mbox{ and S}= 0.0149,  \nonumber \\
&  \mbox{P3 has I}=(0.97)(0.2560) + 0.03 = 0.2783 \mbox{ and S}= 0.7217,  \nonumber \\
&  \mbox{P4 has I}=(0.95)(0.2560) + 0.05 = 0.2932 \mbox{ and S}= 0.7068.  \nonumber
\end{flalign}

\subsubsection{A second example involving four people}

In this example we approach the situation of four infected persons, consider the interactions between three individuals with partial infection:
\begin{flalign}
&  \mbox{P1 has I}=0.01 \mbox{ and S}=0.99, \nonumber \\
&  \mbox{P2 has I}=0.98 \mbox{ and S}=0.02, \nonumber \\
&  \mbox{P3 has I}=0.97 \mbox{ and S}=0.03, \nonumber \\
&  \mbox{P4 has I}=0.99 \mbox{ and S}=0.01. \nonumber 
\end{flalign}
\noindent Once again we  consider contributions from the following: SIII, ISII, IISI, IIIS: 
\begin{flalign}
& \mbox{SIII}=[(0.99)(0.98)(48) + (0.99)(0.97)(36) +  (0.99)(0.99)(64)]/256 = 0.5708,\nonumber \\
& \mbox{ISII}=[(0.02)(0.01)(36) + (0.02)(0.97)(48) +  (0.02)(0.99)(64)]/256 = 0.0086,\nonumber \\
& \mbox{IISI}=[(0.03)(0.01)(36) + (0.03)(0.98)(48) +  (0.03)(0.99)(64)]/256 = 0.0519,\nonumber\\
& \mbox{IIIS}=[(0.01)(0.01)(36) + (0.01)(0.98)(64) +  (0.01)(0.97)(48)]/256 = 0.0043.\nonumber
\end{flalign}
Note from previous examples that if we simply identify the participant with the highest component of $S$ then that will be the one that indicates the highest contribution.  In this case that is P1 and so SIII should be highest, and it is and is 0.5708. This number is very close to  $p_3 = 1 - (3/4)^3 = 0.5781$. It is a higher value because all participants contribute a significant level of infection.  Now we compute the new partial infections for our four participants:
\begin{flalign}
&  \mbox{P1 has I}=(0.99)(0.5708) + 0.01 = 0.5751 \mbox{ and S}= 0.4249,  \nonumber \\
&  \mbox{P2 has I}=(0.02)(0.5708) + 0.98 = 0.9914 \mbox{ and S}= 0.0086,  \nonumber \\
&  \mbox{P3 has I}=(0.03)(0.5708) + 0.97 = 0.9871 \mbox{ and S}= 0.0129,  \nonumber \\
&  \mbox{P4 has I}=(0.01)(0.5708) + 0.99 = 0.9957 \mbox{ and S}= 0.0043.  \nonumber
\end{flalign}

\subsubsection{Algorithm for modelling partially infected people}

For $n+1$ partially infected people meeting at $s$ possible locations, e.g., $s=4$, there are $n$ multiplicative constants. In the first three examples with $n=2$ they are: 16 and 12 which sum to 28.  In the last two examples with $n=3$ they are: 64, 48 and 36 which sum to 148. For $n=1$ there is only one constant: 4.  The constants then get divided by $s^n$ (16, 64, 256,...).

\begin{figure}
\begin{centering}
\includegraphics[width=14.0cm]{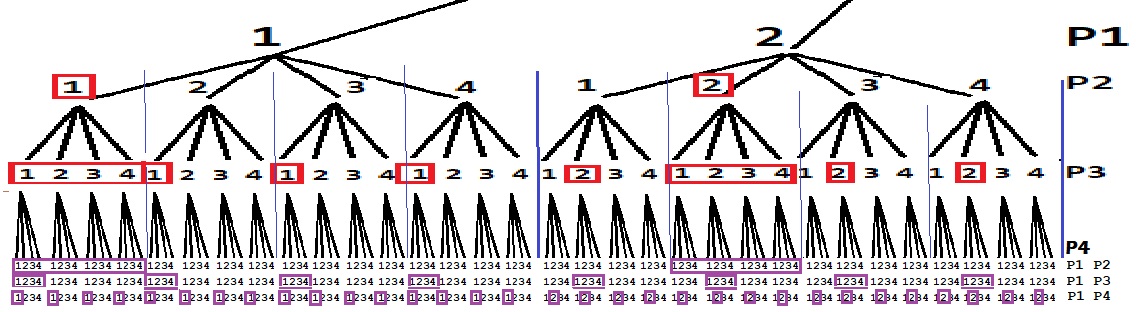}\\
\caption{Location probability tree involving four individuals P1, P2, P3 and $P4$ who may occupy one of four possible locations: 1, 2, 3 or 4.} \label{figlocations2}
\end{centering}
\end{figure}

\begin{figure}
\begin{centering}
\includegraphics[width=5.5cm]{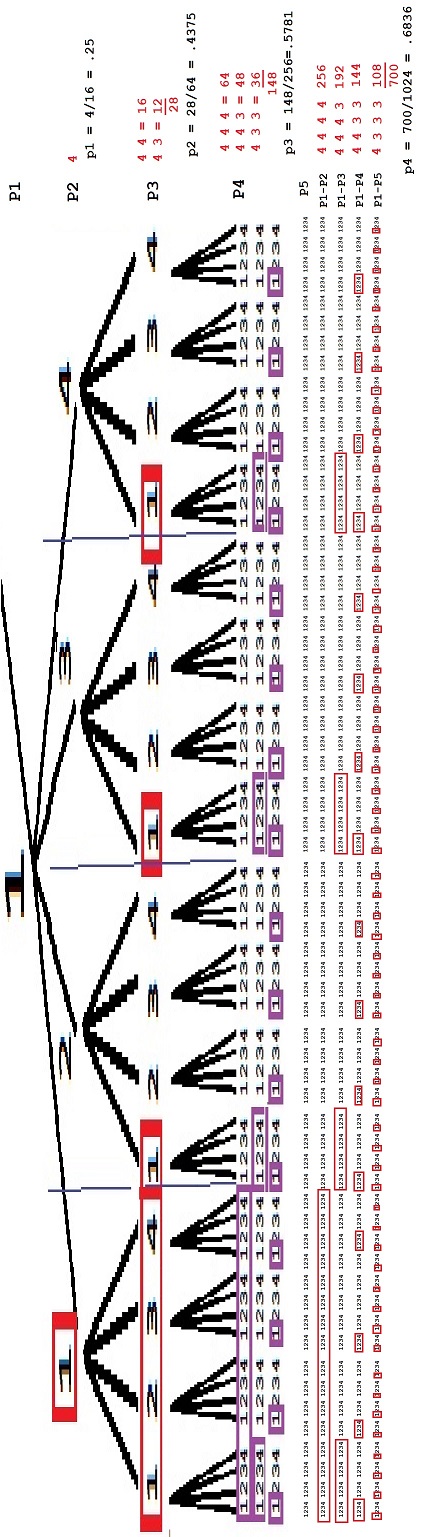}\\
\caption{Understanding the formula for the multiplicative constants pictorially.} \label{figMASSIVE}
\end{centering}
\end{figure}

Figures~\ref{figlocations1} and \ref{figlocations2} illustrate that the sum of these constants divided by $s^n$ gives the infection probability for the pure case of encounter of a fully susceptible individual with $n$ infected, e.g., $p_2 = 28/64 = 0.4375 = 1- (3/4)^2$ and $p_3 = 148/256 = 0.5781 = 1 - (3/4)^3$ . With $n=1$  then $p_1 = 1/4 = 0.25 = 1 - (3/4)$ and the only consideration in such a case are cases SI and IS for the P1-P2 encounter. 
A formula to obtain the $n$ multiplicative constants $k_j$ where $j=1,...,n$ is:
\begin{flalign}
& k_j = s^{n+1-j}  (s-1)^{j-1}....  j=1,...,n. \nonumber
\end{flalign}
This can be understood pictorially at the top (right) of Figure~\ref{figMASSIVE}. When division by $s^n$ is considered the multiplicative ratios $g_j$ in the above examples such as 16/64, 48/256 are given by:
\begin{flalign}
& g_j = (1/s)((s-1)/s)^{j-1}...  j=1,...,n. \nonumber
\end{flalign}
These factors can be seen to be in descending order.  If $n_p$ people participate in an encounter but the number of partially infected people $n$ is less than $n_p$ then we assume one fully susceptible individual with $I=0$ and $S=1$ will meet with the $n$ partially infected individuals and thus $S=1$ is used in the formula.  The calculation follows this ten step procedure (pseudo-code):
\begin{flalign}
&  \mbox{1. pre-compute a large number of factors $g_j$ at the start of the run.} \nonumber \\
&  \mbox{Arriving at each visit $v$ at a location and time:} \nonumber \\
&  \mbox{2. identify all participants $n_p$  that are assigned to this visit $v$};\nonumber \\
&  \mbox{3. if all $n_p$ are fully infected then exit the procedure};\nonumber \\
&  \mbox{4. if none are partially or fully infected exit the procedure};  \nonumber \\
&  \mbox{5. identify the $n$ subset of infected or partially infected out of the $n_p$};  \nonumber \\
&  \mbox{6. if $n<n_p$ include an $(S_v=1$, $I_v=0)$ participant, use $S_{max} = 1$, and increment $n$};  \nonumber \\
&  \mbox{7. otherwise identify $S_{max}$ and that participant with the highest value of $S_v$};  \nonumber \\
&  \mbox{8. sort the infected in descending order by their component of infection $I_v$};  \nonumber \\
&  \mbox{9. prepare the products ($S_{max} I_{vj} g_j$) and sum them to get the infection level};  \nonumber \\
&  \mbox{10. use it to update infection levels of all $n_p$ participants: $I_{v+1}= p S_v + I_v$}.  \nonumber 
\end{flalign}


\newpage
\newpage
\begin{figure}
\begin{centering}
\includegraphics[width=12.0cm]{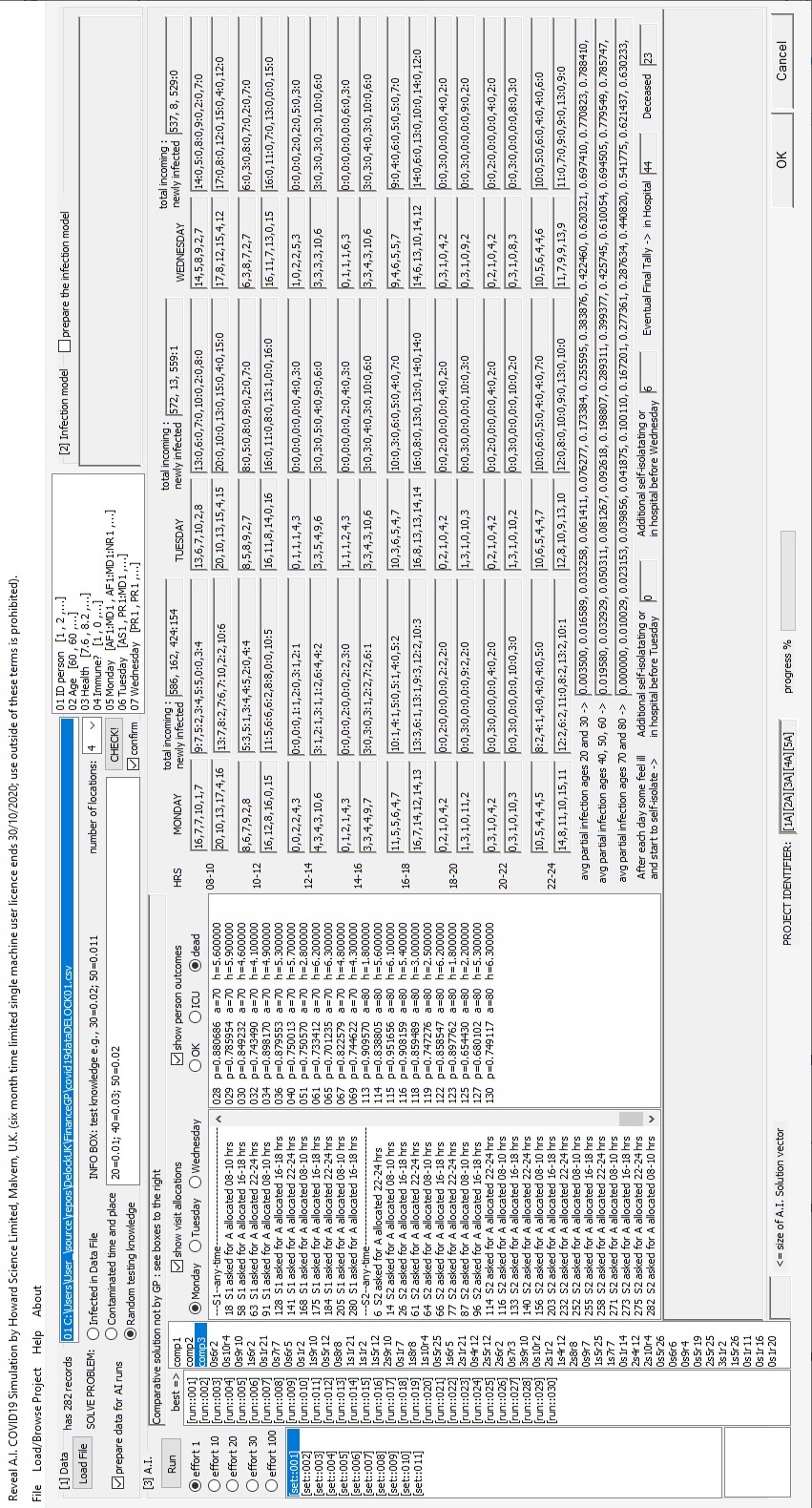}\\
\caption{ $comp3$ result: 20-year-olds 1 \% infected, 40-year-olds 3 \% infected and 50-year-olds 2\%  infected, with $s=4$.} \label{fig:pI03}
\end{centering}
\end{figure}
\newpage
\begin{figure}
\begin{centering}
\includegraphics[width=12.0cm]{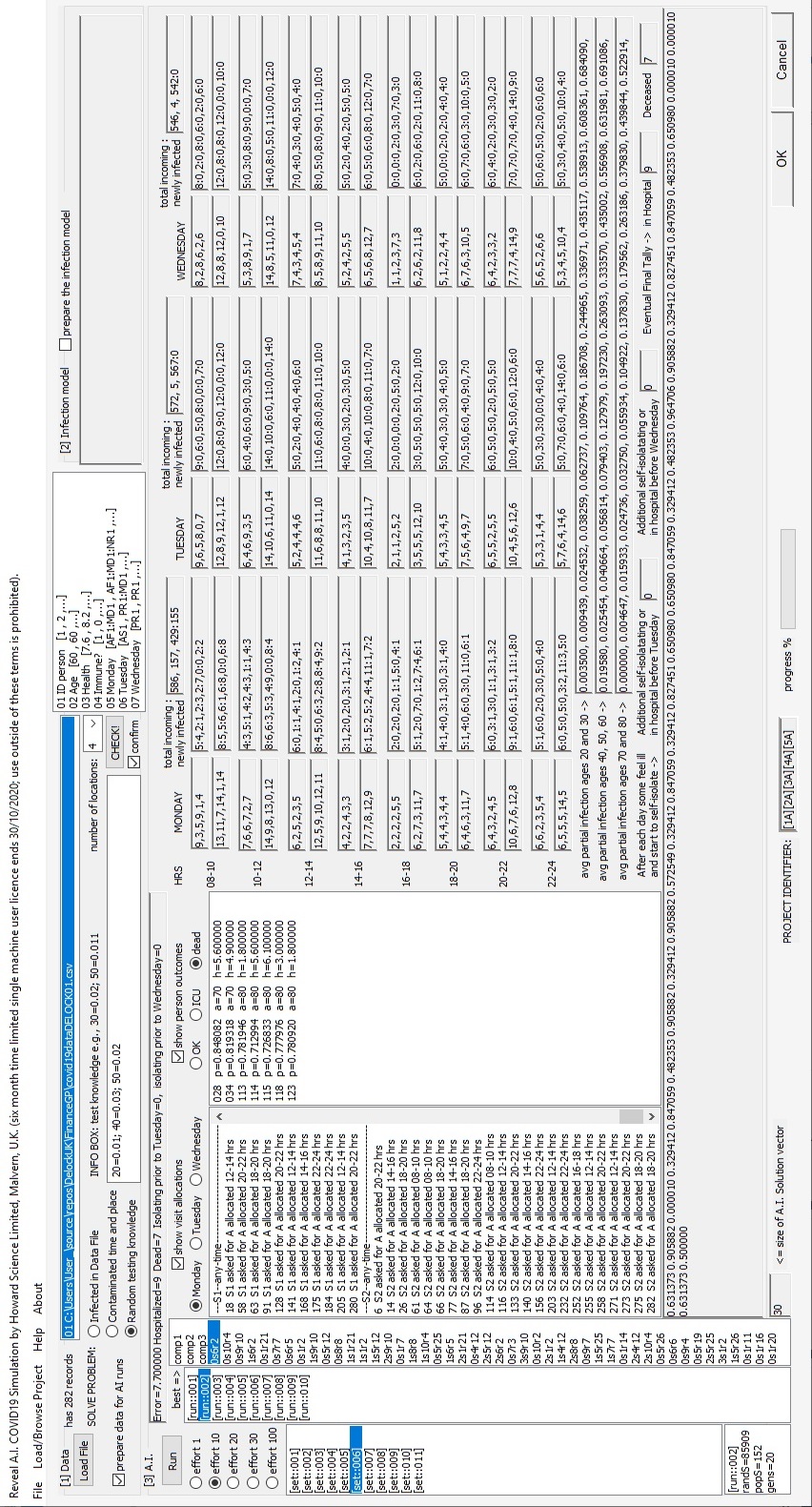}\\
\caption{ GP result: 20-year-olds 1 \% infected, 40-year-olds 3 \% infected and 50-year-olds 2\%  infected, with $s=4$.} \label{fig:pI04}
\end{centering}
\end{figure}
\newpage
\begin{figure}
\begin{centering}
\includegraphics[width=12.0cm]{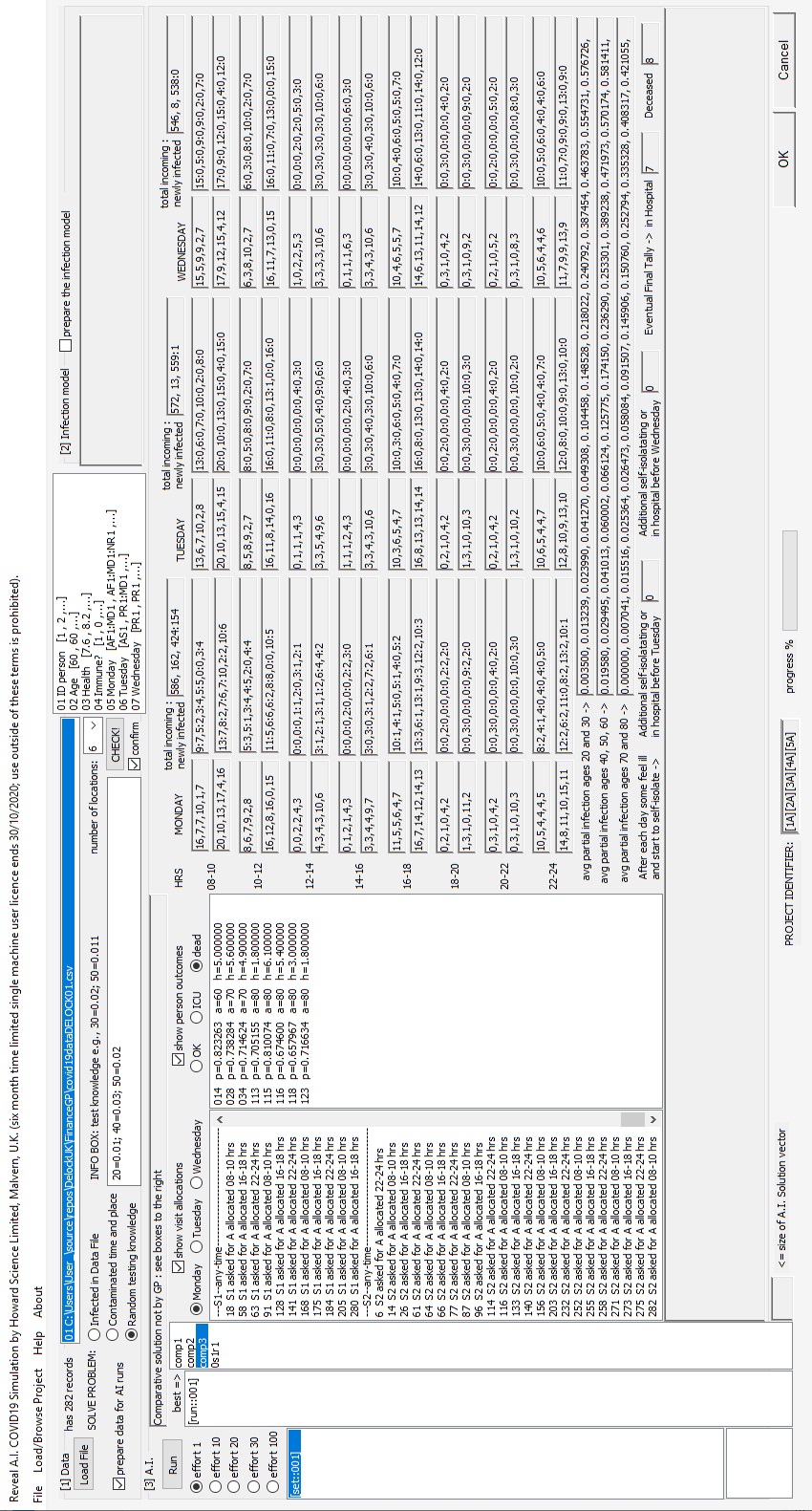}\\
\caption{ $comp3$ result: 20-year-olds 1 \% infected, 40-year-olds 3 \% infected and 50-year-olds 2\%  infected, with $s=6$.} \label{fig:pI07}
\end{centering}
\end{figure}
\newpage
\begin{figure}
\begin{centering}
\includegraphics[width=12.0cm]{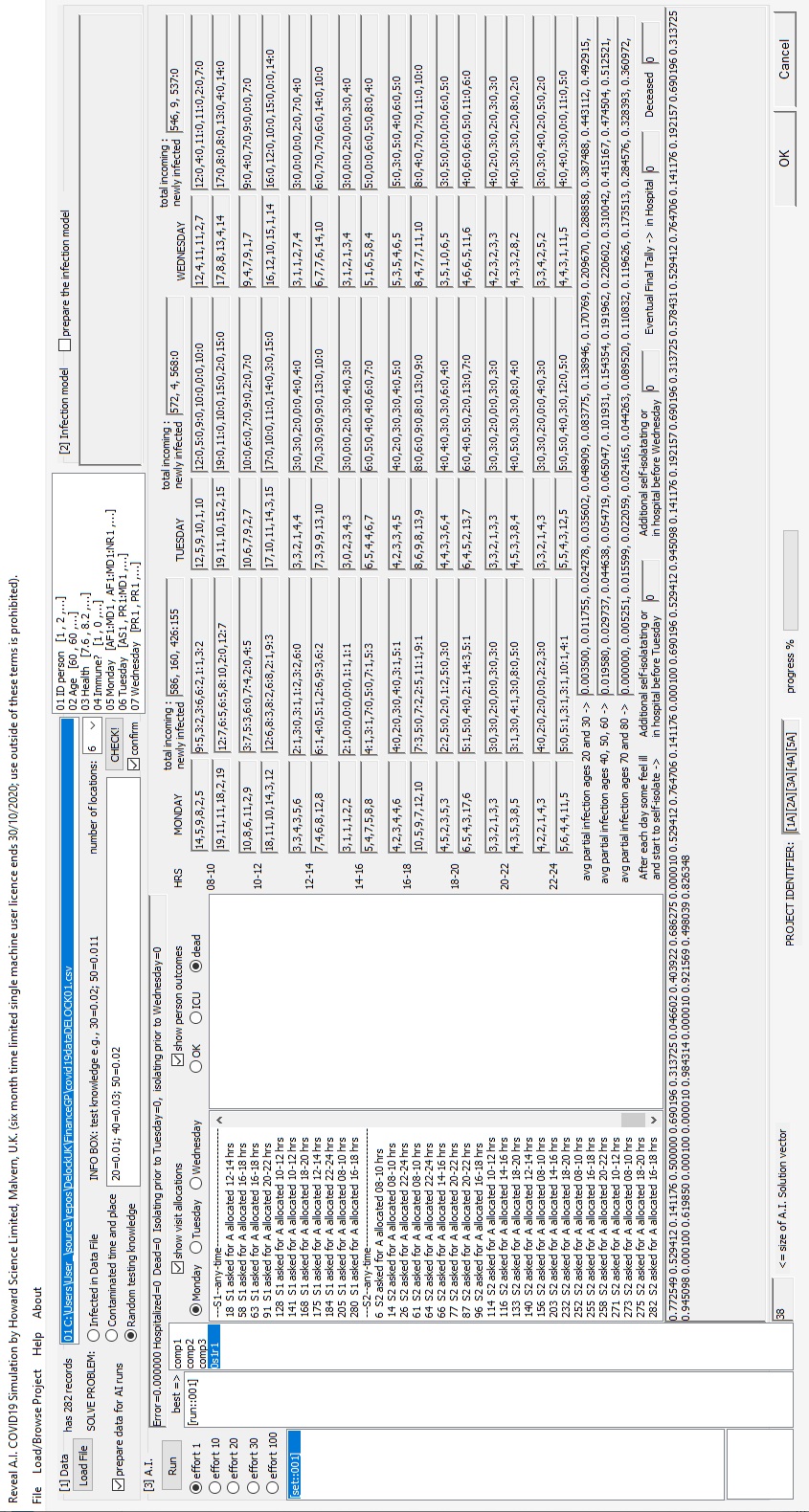}\\
\caption{ GP result: 20-year-olds 1 \% infected, 40-year-olds 3 \% infected and 50-year-olds 2\%  infected, with $s=6$.} \label{fig:pI08}
\end{centering}
\end{figure}
\newpage
\begin{figure}
\begin{centering}
\includegraphics[width=12.0cm]{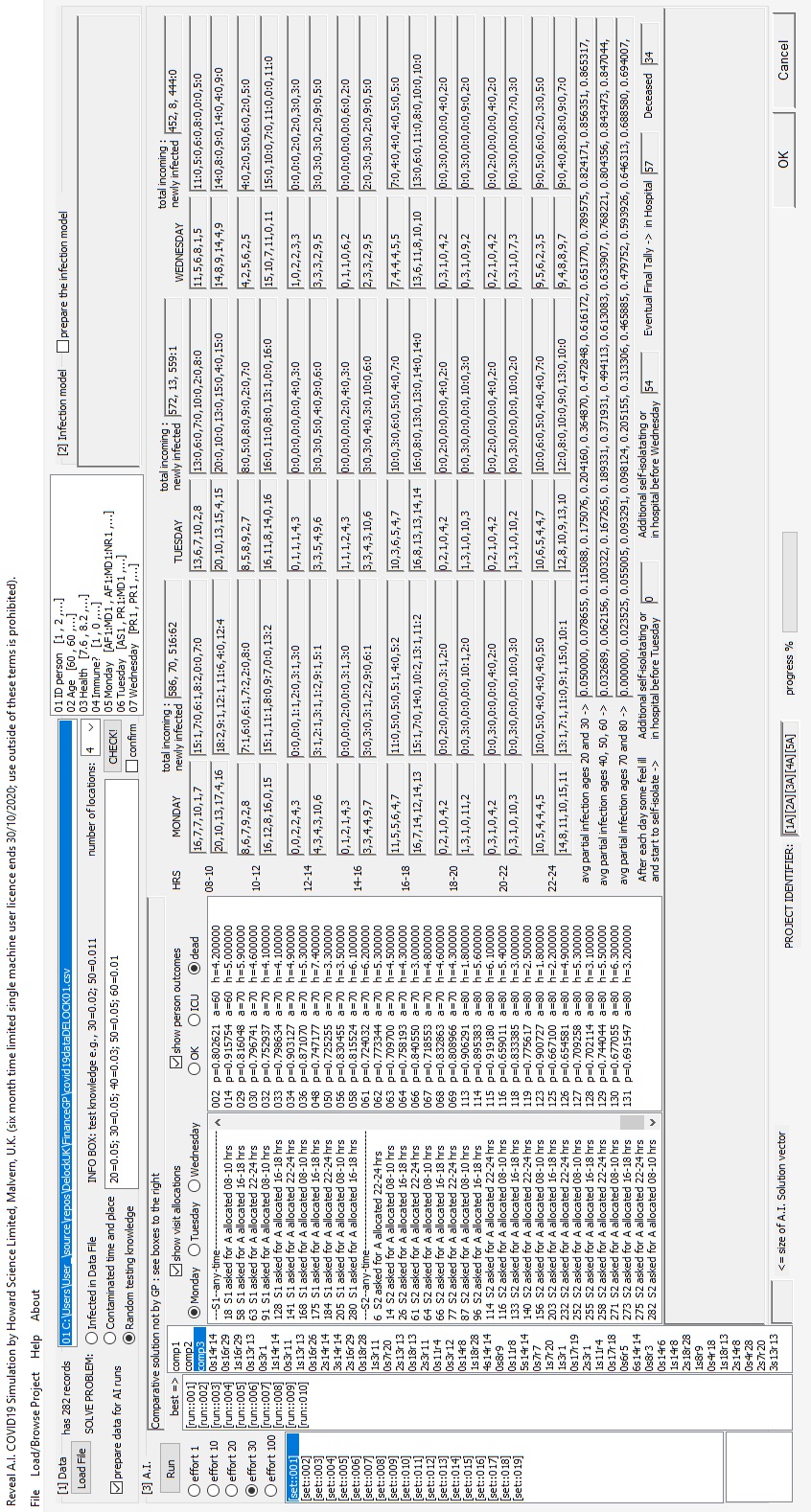}\\
\caption{ $comp3$ result: 20-year-olds 5 \% infected, 30-year-olds 5 \% infected, 40-year-olds 3 \% infected, 50-year-olds 5\%  infected and 60-year-olds 1 \%infected, with $s=4$.} \label{fig:pI11}
\end{centering}
\end{figure}
\newpage
\begin{figure}
\begin{centering}
\includegraphics[width=12.0cm]{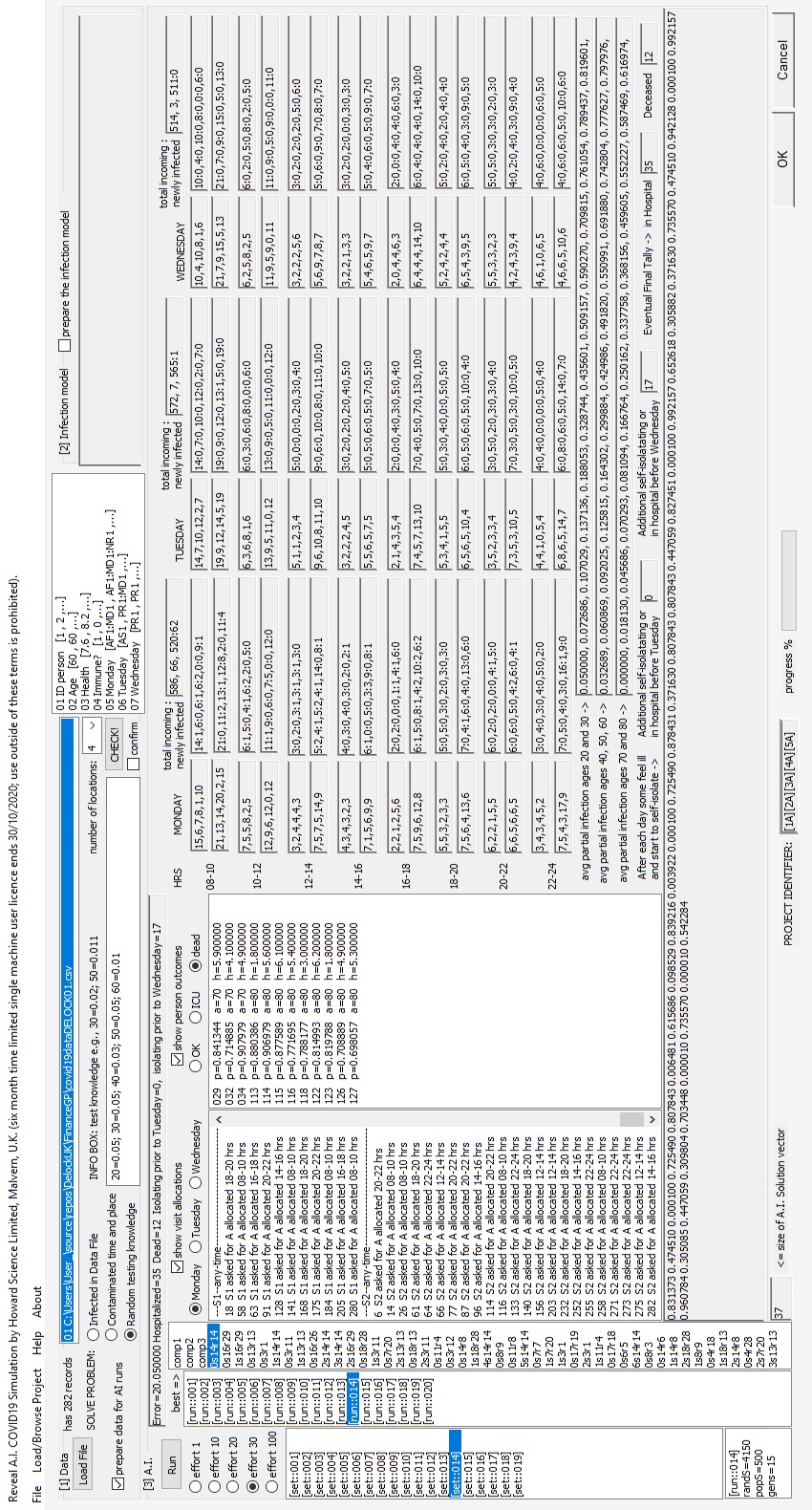}\\
\caption{ GP result: 20-year-olds 5 \% infected, 30-year-olds 5 \% infected, 40-year-olds 3 \% infected, 50-year-olds 5\%  infected and 60-year-olds 1 \%infected, with $s=4$.} \label{fig:pI12}
\end{centering}
\end{figure}
\newpage
\begin{figure}
\begin{centering}
\includegraphics[width=12.0cm]{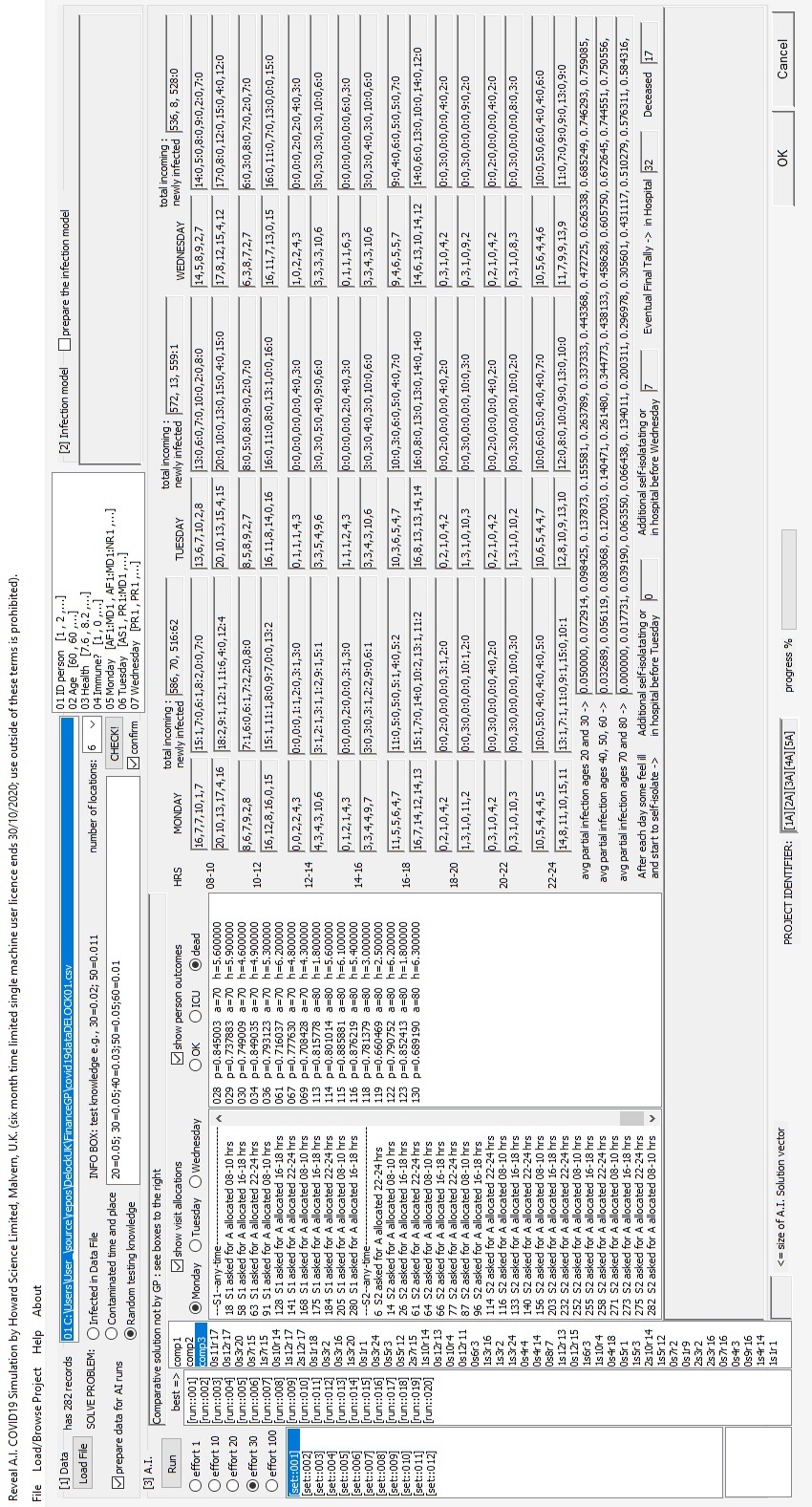}\\
\caption{ $comp3$ result: 20-year-olds 5 \% infected, 30-year-olds 5 \% infected, 40-year-olds 3 \% infected, 50-year-olds 5\%  infected and 60-year-olds 1 \%infected, with $s=6$.} \label{fig:pI16}
\end{centering}
\end{figure}
\newpage
\begin{figure}
\begin{centering}
\includegraphics[width=12.0cm]{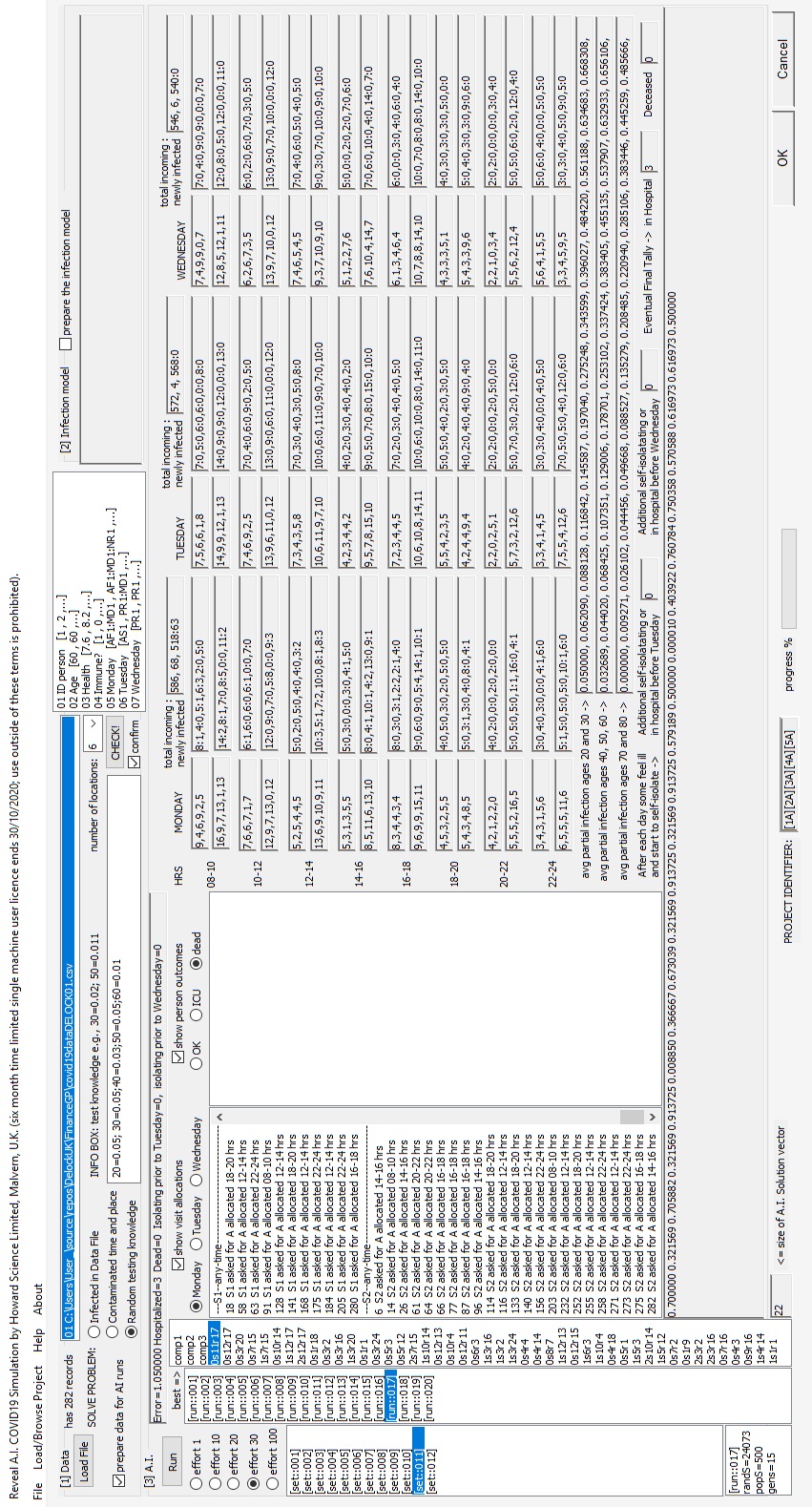}\\
\caption{ GP result: 20-year-olds 5 \% infected, 30-year-olds 5 \% infected, 40-year-olds 3 \% infected, 50-year-olds 5\%  infected and 60-year-olds 1 \%infected, with $s=6$.} \label{fig:pI17}
\end{centering}
\end{figure}
\newpage
\begin{figure}
\begin{centering}
\includegraphics[width=12.0cm]{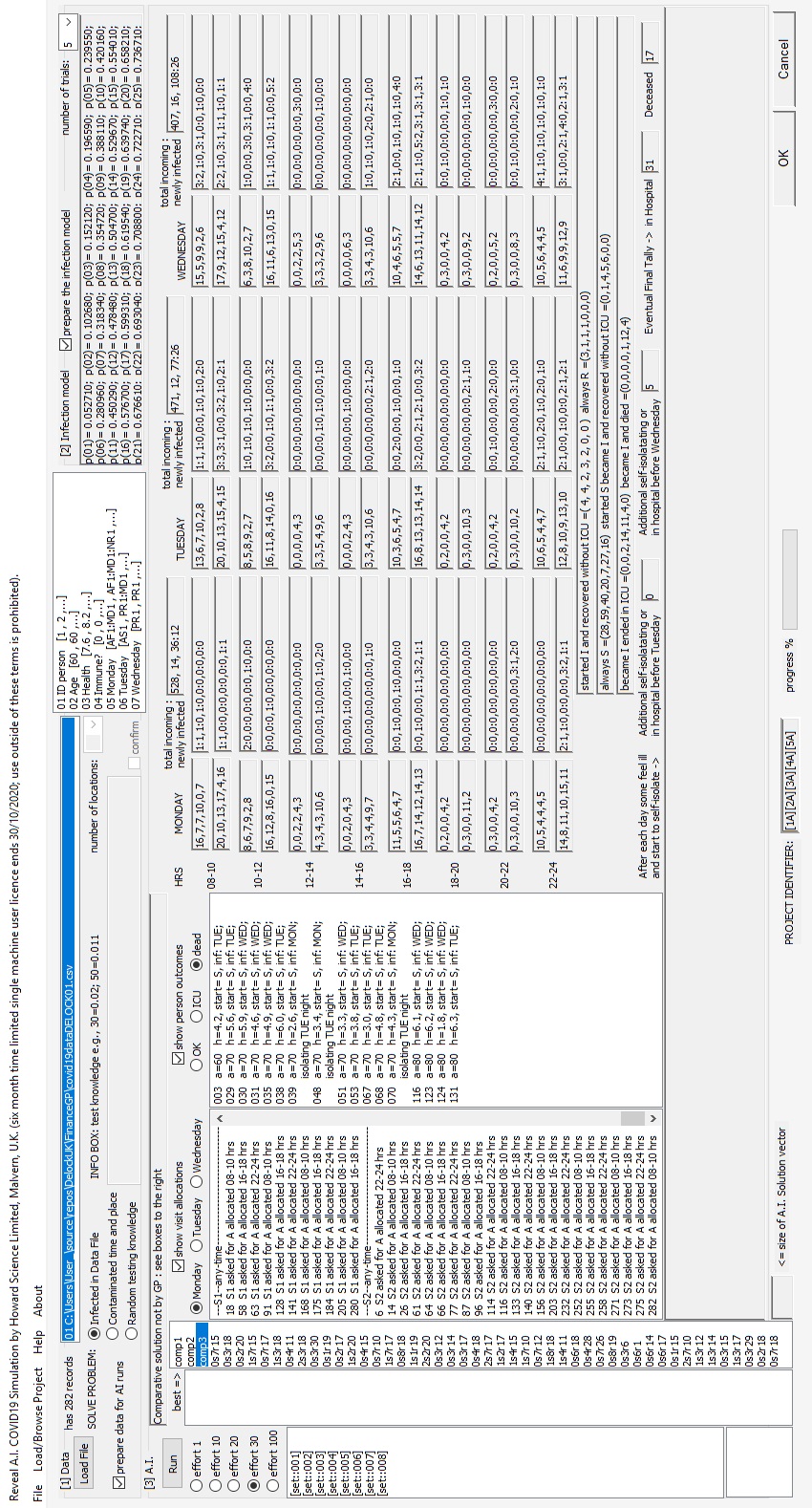}\\
\caption{$comp3$: full infection with $q=5$ in Monte Carlo probability of infection procedure - see Appendix~\ref{appendix1}.} \label{fig:full14}
\end{centering}
\end{figure}
\newpage
\begin{figure}
\begin{centering}
\includegraphics[width=12.0cm]{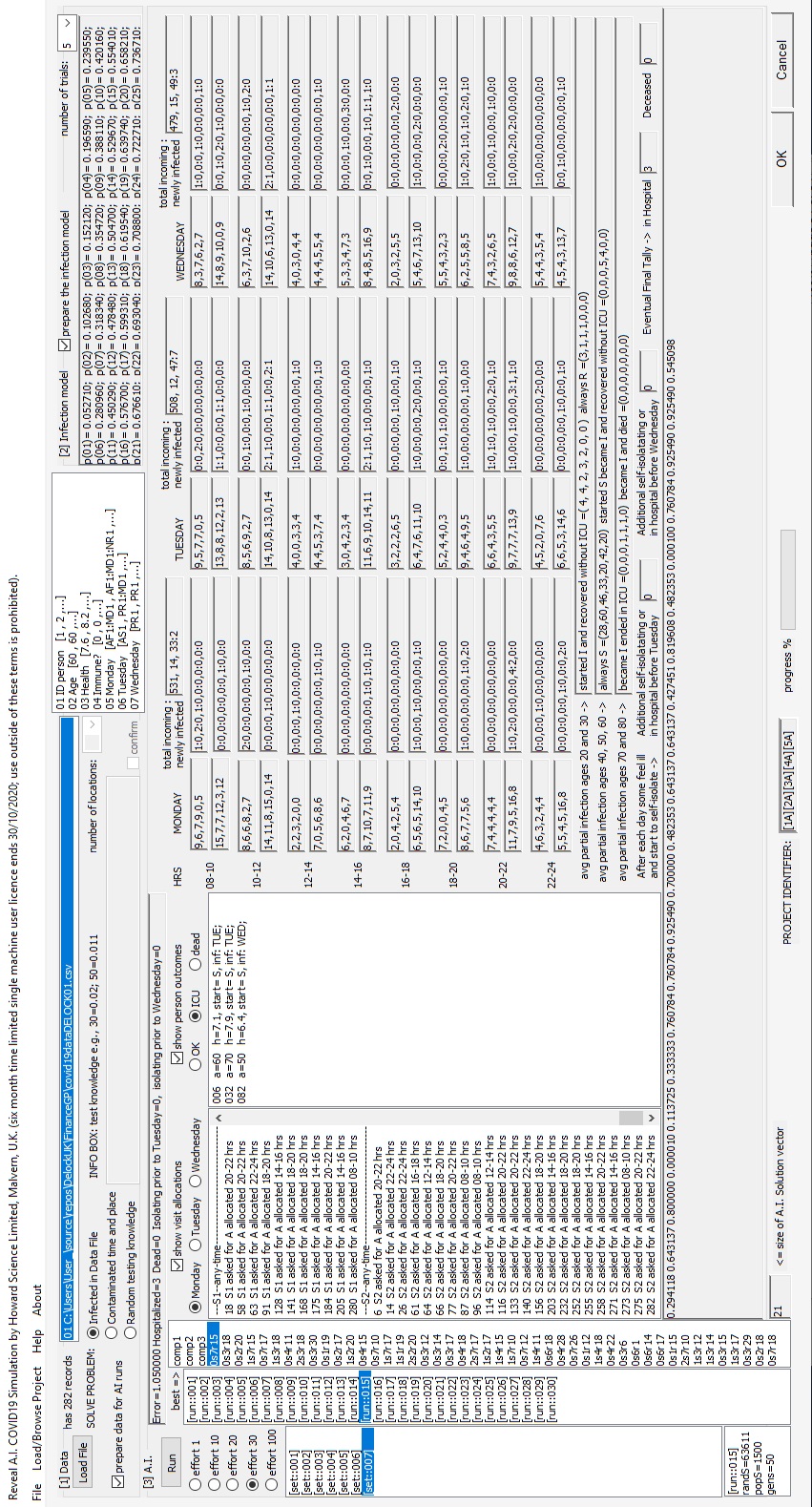}\\
\caption{ GP result: full infection with $q=5$ in Monte Carlo probability of infection procedure - see Appendix~\ref{appendix1}.} \label{fig:full15}
\end{centering}
\end{figure}
\newpage
\begin{figure}
\begin{centering}
\includegraphics[width=12.0cm]{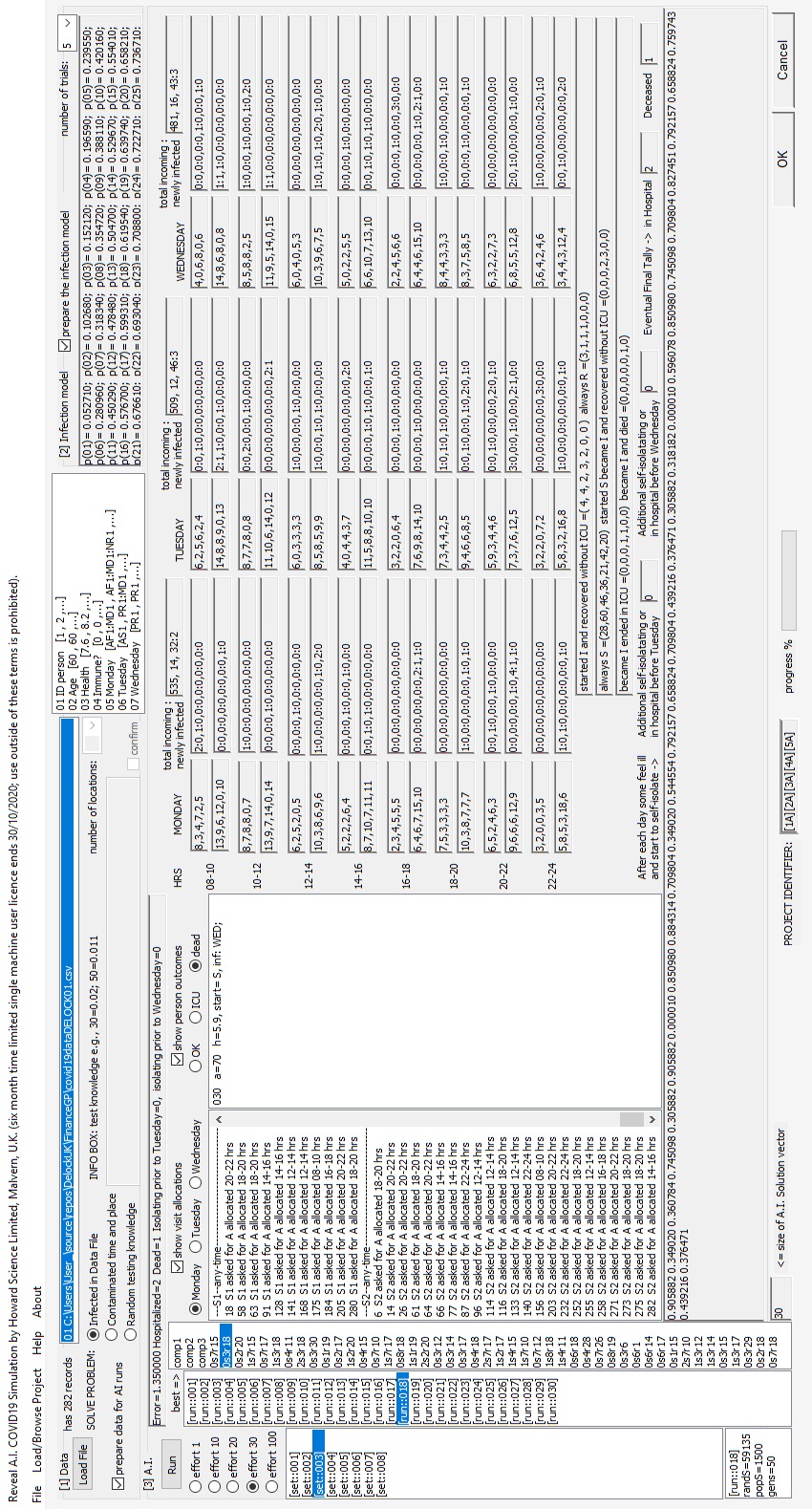}\\
\caption{ GP result: full infection with $q=5$ in Monte Carlo probability of infection procedure - see Appendix~\ref{appendix1}.} \label{fig:full16}
\end{centering}
\end{figure}
\newpage
\begin{figure}
\begin{centering}
\includegraphics[width=12.0cm]{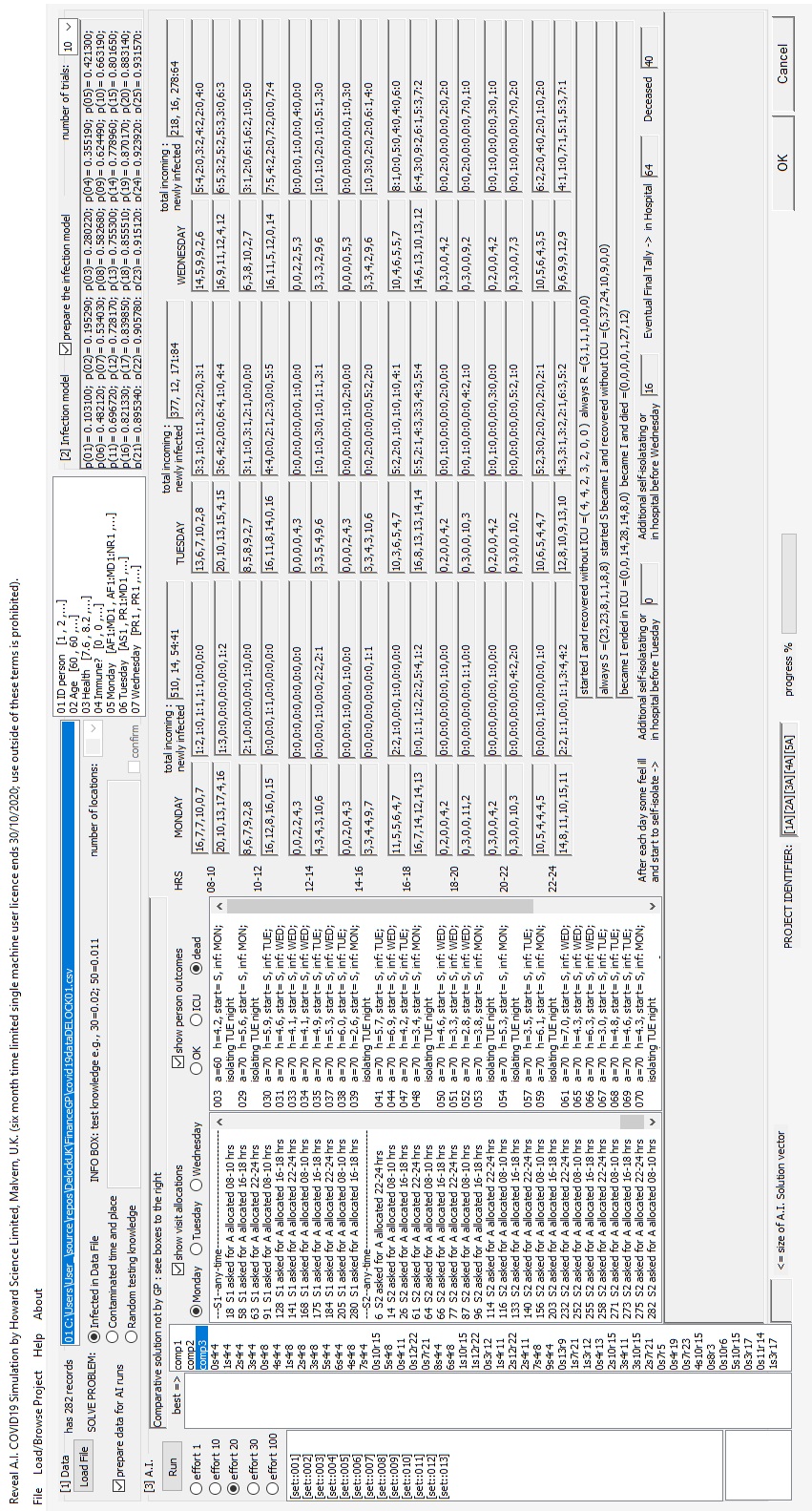}\\
\caption{$comp3$: full infection with $q=10$ in Monte Carlo probability of infection procedure - see Appendix~\ref{appendix1}.} \label{fig:full03}
\end{centering}
\end{figure}
\newpage
\begin{figure}
\begin{centering}
\includegraphics[width=12.0cm]{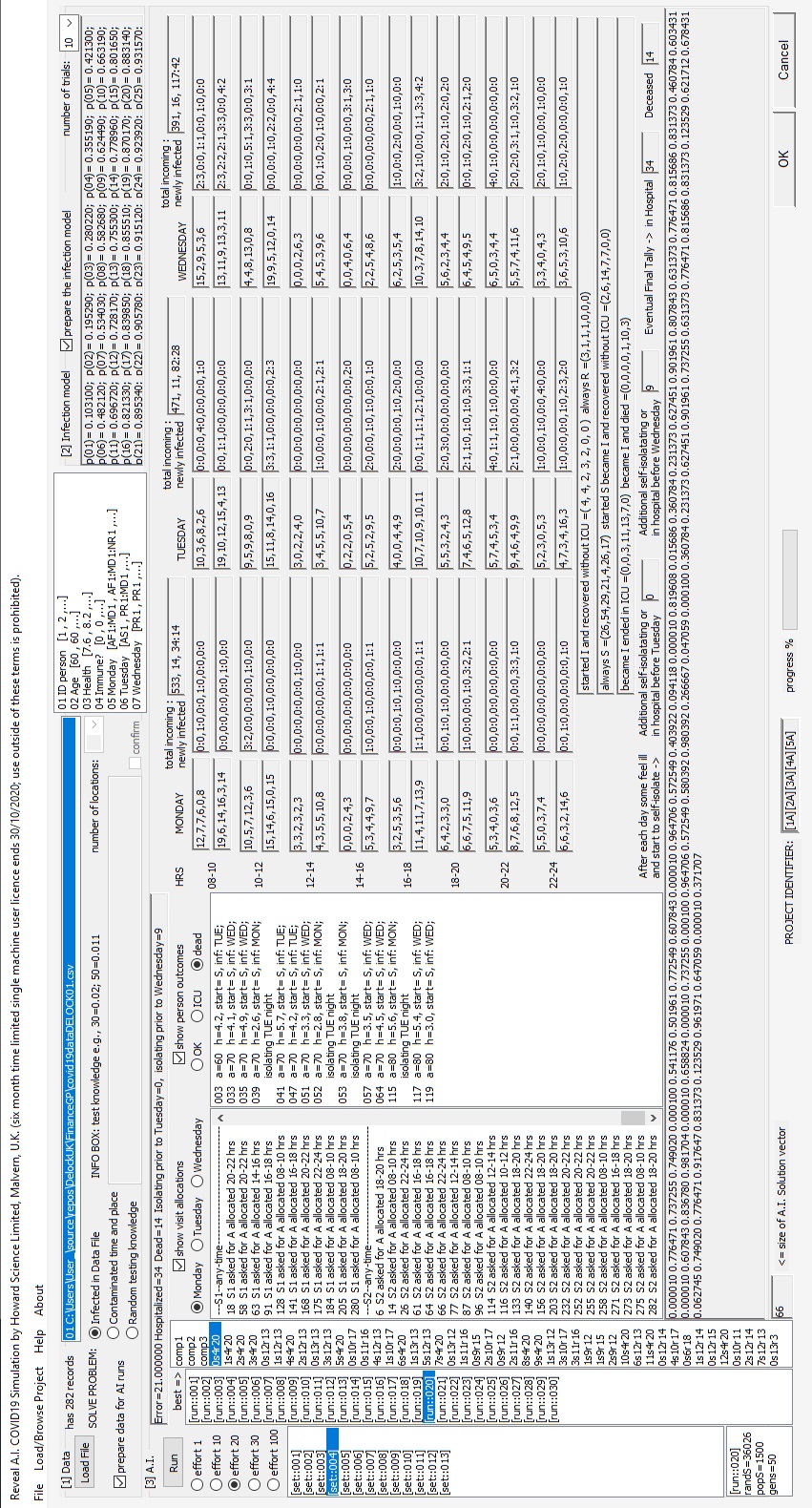}\\
\caption{ GP result: full infection with $q=10$ in Monte Carlo probability of infection procedure - see Appendix~\ref{appendix1}.} \label{fig:full04}
\end{centering}
\end{figure}
\newpage
\begin{figure}
\begin{centering}
\includegraphics[width=12.0cm]{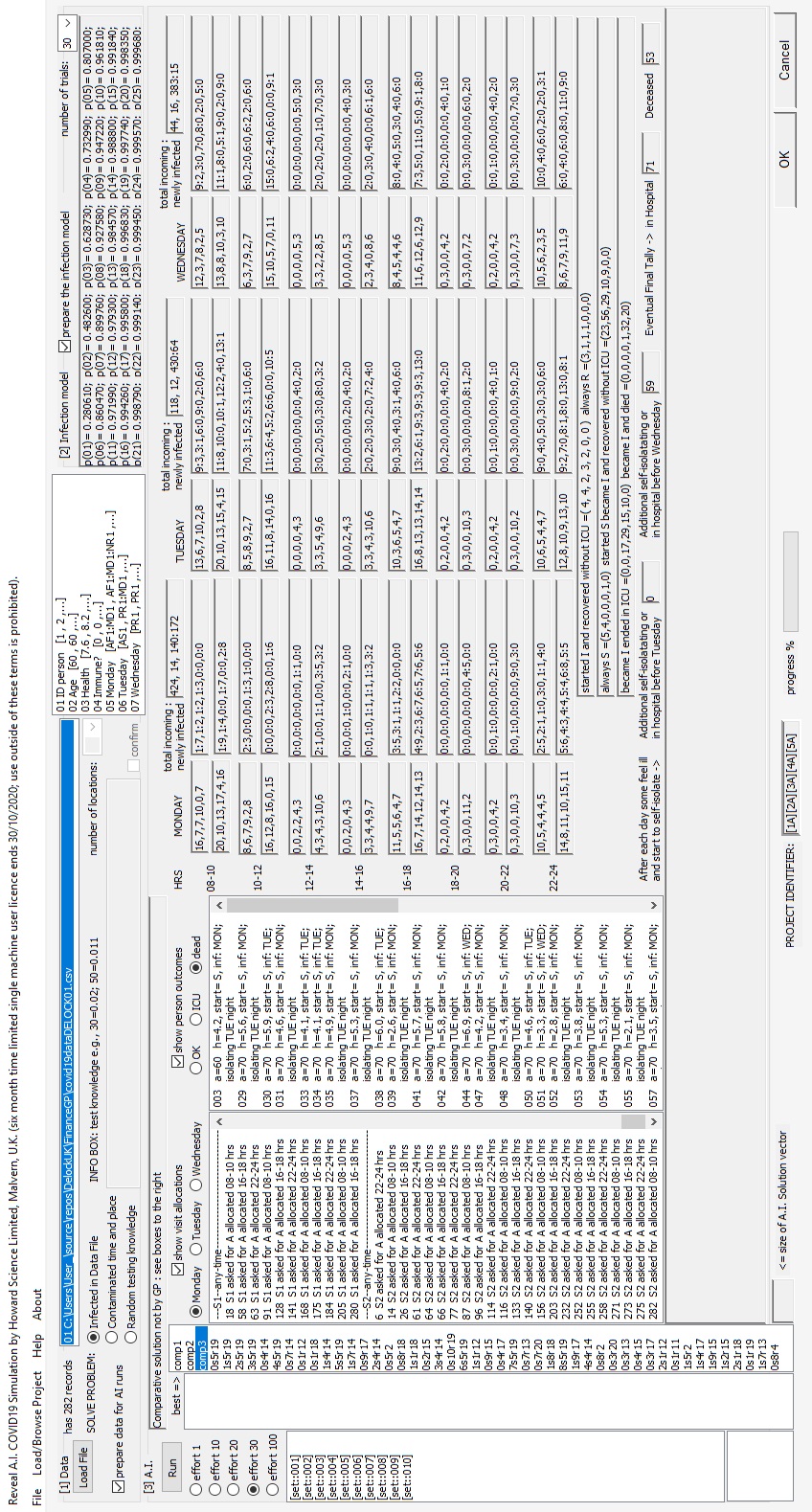}\\
\caption{$comp3$: full infection with $q=30$ in Monte Carlo probability of infection procedure - see Appendix~\ref{appendix1}.} \label{fig:full09}
\end{centering}
\end{figure}
\newpage
\begin{figure}
\begin{centering}
\includegraphics[width=12.0cm]{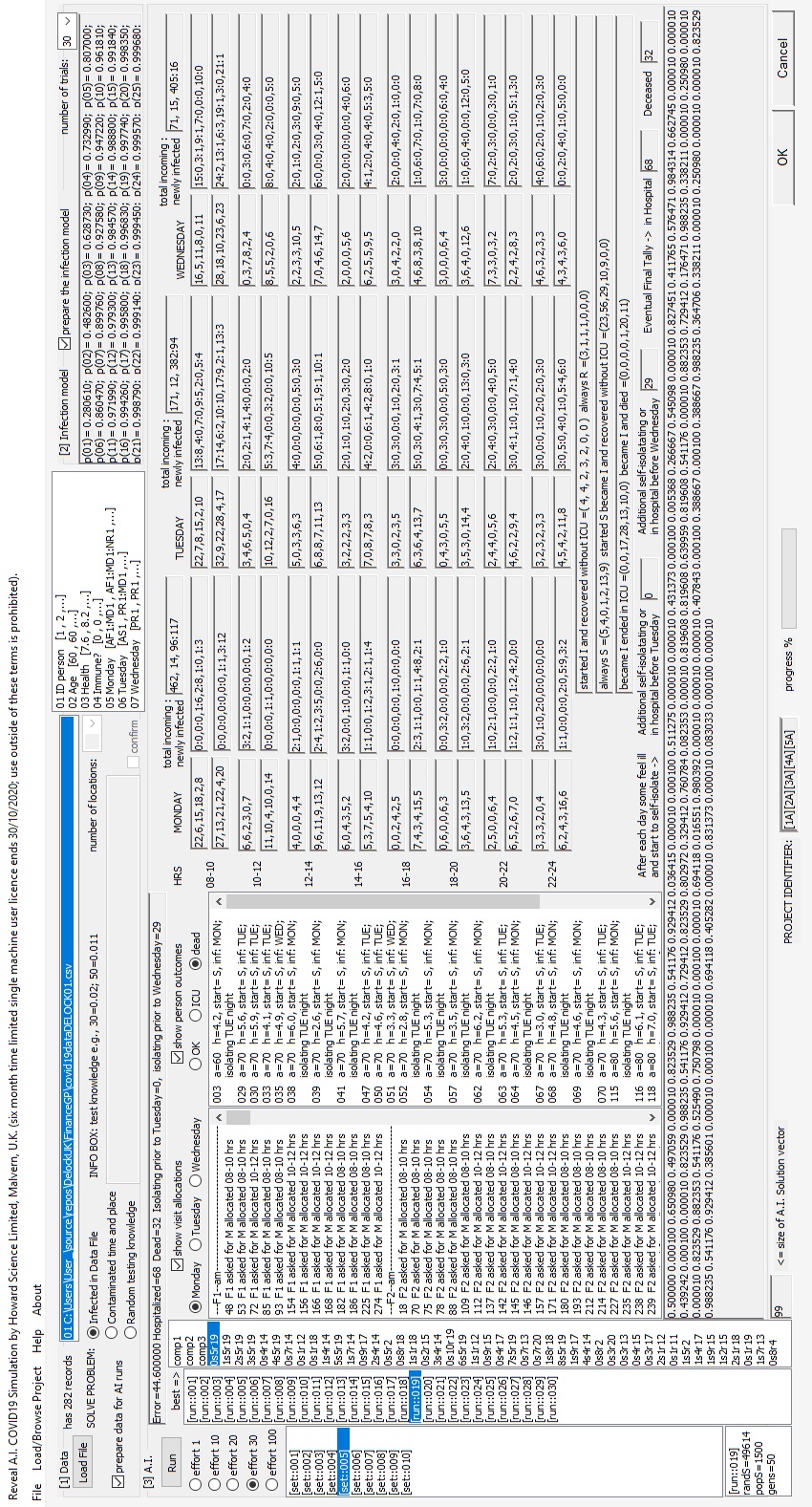}\\
\caption{ GP result: full infection with $q=30$ in Monte Carlo probability of infection procedure - see Appendix~\ref{appendix1}.} \label{fig:full10}
\end{centering}
\end{figure}
\newpage

\bibliographystyle{plain}
\bibliography{arXivDH1}

\end{document}